\documentclass[preprint,12pt,authoryear]{elsarticle}

\usepackage{amssymb}
\usepackage[ruled]{algorithm2e}
\usepackage{setspace}
\usepackage{amsmath}
\usepackage{longtable}
\usepackage{float}
\usepackage{graphicx}
\usepackage{threeparttable}
\usepackage[caption=false,font=small,labelfon
t=sf,textfont=sf]{subfig}
\usepackage{hyperref}
\usepackage{siunitx}
\usepackage{verbatim}
\usepackage[utf8]{inputenc}
\usepackage{bbm}
\usepackage{color}
\usepackage{xcolor}
\usepackage{threeparttable}
\usepackage{makecell}
\usepackage{multirow}

\journal{Transportation Research Part C}

\begin{document}

\begin{frontmatter}



\title{Modified DDPG car-following model with a real-world human driving experience with CARLA simulator}

\author[inst1]{Dianzhao Li\corref{cor1}}
\ead{dianzhao.li@tu-dresden.de}
\author[inst1]{Ostap Okhrin}
\affiliation[inst1]{organization={Chair of Econometrics and Statistics, esp. in the Transport Sector, \\
Technische Universität Dresden},
            city={Dresden},
            country={Germany}}
\cortext[cor1]{Corresponding author}



\begin{abstract}
In the autonomous driving field, fusion of human knowledge into Deep Reinforcement Learning (DRL) is often based on the human demonstration recorded in a simulated environment. This limits the generalization and the feasibility of application in real-world traffic. We propose a two-stage DRL method to train a car-following agent, that modifies the policy by leveraging the real-world human driving experience and achieves performance superior to the pure DRL agent. Training a DRL agent is done within CARLA framework with Robot Operating System (ROS). For evaluation, we designed different driving scenarios to compare the proposed two-stage DRL car-following agent with other agents. After extracting the "good" behavior from the human driver, the agent becomes more efficient and reasonable, which makes this autonomous agent more suitable to Human-Robot Interaction (HRI) traffic.

\end{abstract}




\end{frontmatter}


\section{Introduction}
\label{sec:Introduction}

With the vigorous development of new energy vehicles, autonomous driving has become more and more attractive to the academia. Due to the complex real-world traffic environment, vehicles are making continuously decisions that are building cornerstones in achieving fully autonomous driving. A list of solutions for complex decision-making problems has been proposed in the literature over the last decades.\par 

The first family of autonomous techniques is the so-called \textit{rule-based} control strategies, which use lists of complex control rules to define the behavior of vehicles in the traffic flow. For the car-following task in autonomous driving, models worth mentioning are the Gaxis-Herman-Rothery (GHR) model by \citet{gazis1961nonlinear}, the \citet{Wiedemann.1974} car-following model, Intelligent Driver Model (IDM) by \cite{treiber2000congested} or stochastic car-following models by \cite{treiber2017intelligent}. 
\citet{kikuchi1992car} proposed a fuzzy inference system-based car-following model, which consists of many direct natural language-based driving rules. Obviously, it is impossible to consider all the situations that may occur and formulate corresponding control strategies accordingly. Moreover, rule-based control strategies are not suitable for time-varying and non-stationary traffic conditions. \par

Due to the rapid development of deep learning in recent years, second approach to solve control problems involves the use of \textit{imitation learning} (IL) or \textit{behavior cloning} (BC). With BC, the perception and control parts in autonomous driving are learned from human demonstrations using deep neural networks (DNN). The DNN learns, from an RGB image from the camera or corresponding features obtained from other sensors like LiDAR, Laser, Ultrasonic, and GPS as input and outputs the desired reaction to the vehicle control in certain circumstances. End-to-end imitation systems can be learned offline in a safe way. \cite{bojarski2016end} trained a convolutional neural network (CNN) to map raw images from a single camera directly to steering commands. \cite{codevilla2018end} proposed command-conditional IL, in which low-level controls, as well as high-level commands, are learned from expert demonstrations with vision-based inputs.  \par
                                    
Although BC is widely used, even in real cars, that are allowed for driving on the streets, it still has many shortcomings. \cite{codevilla2019exploring} pointed out four main limitations: 
\begin{itemize}
    \item [] \emph{Generalization}: As a type of supervised learning, BC is also limited by the size of the dataset and for scenarios, not appearing in the dataset, the performance will be unsatisfactory.
    \item [] \emph{Driving dataset biases}: Supervised learning is limited by the type or the source of the dataset, and the trained control strategy will be biased. Diversity is an important criterion for a dataset, as every driver has his own driving style, and even different vehicles have different operation performances, which are reflected in the datasets. 
    
    \item [] \emph{Causal confusion}: Some spurious correlations cannot be distinguished from true causes in human demonstration patterns if we do not use an explicit causal model or on-policy demonstrations, see  \cite{de2019causal}.
    \item [] \emph{High variance}: Since off-policy IL uses a static dataset, initialization and data sampling will induce high variance. As the cost function in BC is optimized via Stochastic Gradient Descent, which assumes the data is independent and identically distributed. However, the human demonstration dataset is usually interrelated over a long period of time. Therefore the model can be very sensitive to the initialization, see \cite{hanin2018start}.
\end{itemize}

A feasible solution to the challenge above that has attracted widespread attention in the academic community over the last decades is \textit{Deep Reinforcement Learning} (DRL). DRL that combines reinforcement learning and deep learning, has provided solutions for many complex decision-making tasks on playing chess, Go, Atari games, uses in robotics, drones, and vehicle behavior (\citealp{mnih2013playing}; \citealp{silver2016mastering}; \citealp{silver2017mastering}; \citealp{gu2017deep}; \citealp{akhloufi2019drones}; \citealp{he2017integrated}; \citealp{isele2018navigating}). 
Within this field, agents interact with the environments to learn the optimal behaviors, improving over time through trial and error, and in general, do not base on the datasets.\par

Although DRL methods are present, unfortunately, are not prominent in transportation scenarios. For example, \cite{sallab2016end} used Deep Q learning (DQN) and Deep Deterministic Actor Critic Algorithm (DDAC) as control algorithms for lane-keeping tasks in TORCS simulator. Later, \citet{wang2018reinforcement} proposed Q-learning based training process to train a control policy for lane-changing tasks with a smooth and efficient behavior. The study by \cite{ngai2011multiple} offered a multiple-objective RL framework, which is used to solve the overtaking problem in traffic flow with correct action as well as collision avoidance. Recently, \cite{nosrati2018towards} proposed an RL-based hierarchical framework with DQN and Proximal Policy Optimization (PPO) to solve the multi-lane cruising problem, showing that this design enables better learning performance. \cite{hart2021formulation} proposed a DRL car-following model which decomposes the multi-goal problem into two subtasks: free-driving and car-following with a Modularized Reinforcement Learning approach,  therefore the car-following agent can follow the leader with the desired speed and also keep a reasonable gap to the leader.\par

In addition to pure DRL, researchers are combining DRL with human prior knowledge to improve exploration efficiency. \cite{hester2018deep} proposed the Deep Q-learning from Demonstration (DQfD) algorithm, which maintained two separate replay buffers to store demonstration data and self-generated data respectively. They showed that DQfD performs better than other related algorithms for incorporating demonstration data into DQN. \cite{vecerik2017leveraging} utilized human demonstrations with Deep Deterministic Policy Gradient (DDPGfD) algorithm. The human demonstrations and actual interactions are filled into the replay buffer and sampled with prioritized replay mechanism. The results on simulated tasks showed that DDPG with demonstrations outperforms pure DDPG. \cite{huang2021efficient} proposed a framework that first uses BC with the expert demonstration to derive an imitative expert policy and then further improve it with DRL. A different approach has been done by \cite{liu2021improved} who modified the update of the policy network in RL to leverage human prior knowledge, which can adaptively sample experience from the agent’s self-exploration and expert demonstration for the policy update. \par

All aforementioned methods combining human demonstrations with DRL has four limitations: First, application of supervised learning to human demonstrations and obtaining a decent initialization for the DRL policy network does not guarantee a more comprehensive utilization of the demonstrations, a pure DRL agent is often unable to safely interact with human drivers in real-world traffic \citep{litman2017autonomous}. Second, training a human-like agent with DRL or BC usually requires fairly large datasets, otherwise, the performance is poor. The real-world datasets are not enough to cover all the possible traffic scenarios. Moreover, in the real-world driving datasets, for instance, NGSIM and HighD datasets, the drivers are influenced by the complex real traffic environment and cannot focus on only one certain driving task. Since we want our agent to focus on the car-following behavior only, the commonly used datasets need to be re-extracted. Therefore the size of the datasets will be reduced greatly. In addition, recording data in the simulation usually requires dozens of hours of operations by human experts, which is very resource-intensive. Third, most of the papers combine human demonstrations with DRL, which are generated and recorded in a simulated environment. This leads to restrictions on generalization and dataset biases issues when using the data (\citealp{huang2021efficient}; \citealp{liu2021improved}). Fourth, while training a DRL agent with human demonstrations, an important issue for this off-policy learning is \textit{extrapolation error} \citep{fujimoto2019off}, which occurs when a mismatch happens between the distribution of data induced by the current policy and the distribution of data contained in the batch. Consequently, it will be impossible to learn a value function for a policy that selects actions not contained in the batch. In the next, we briefly outline our contribution. \par

\subsection{Contribution}

\begin{figure}
    \centering
    \includegraphics[width=5in]{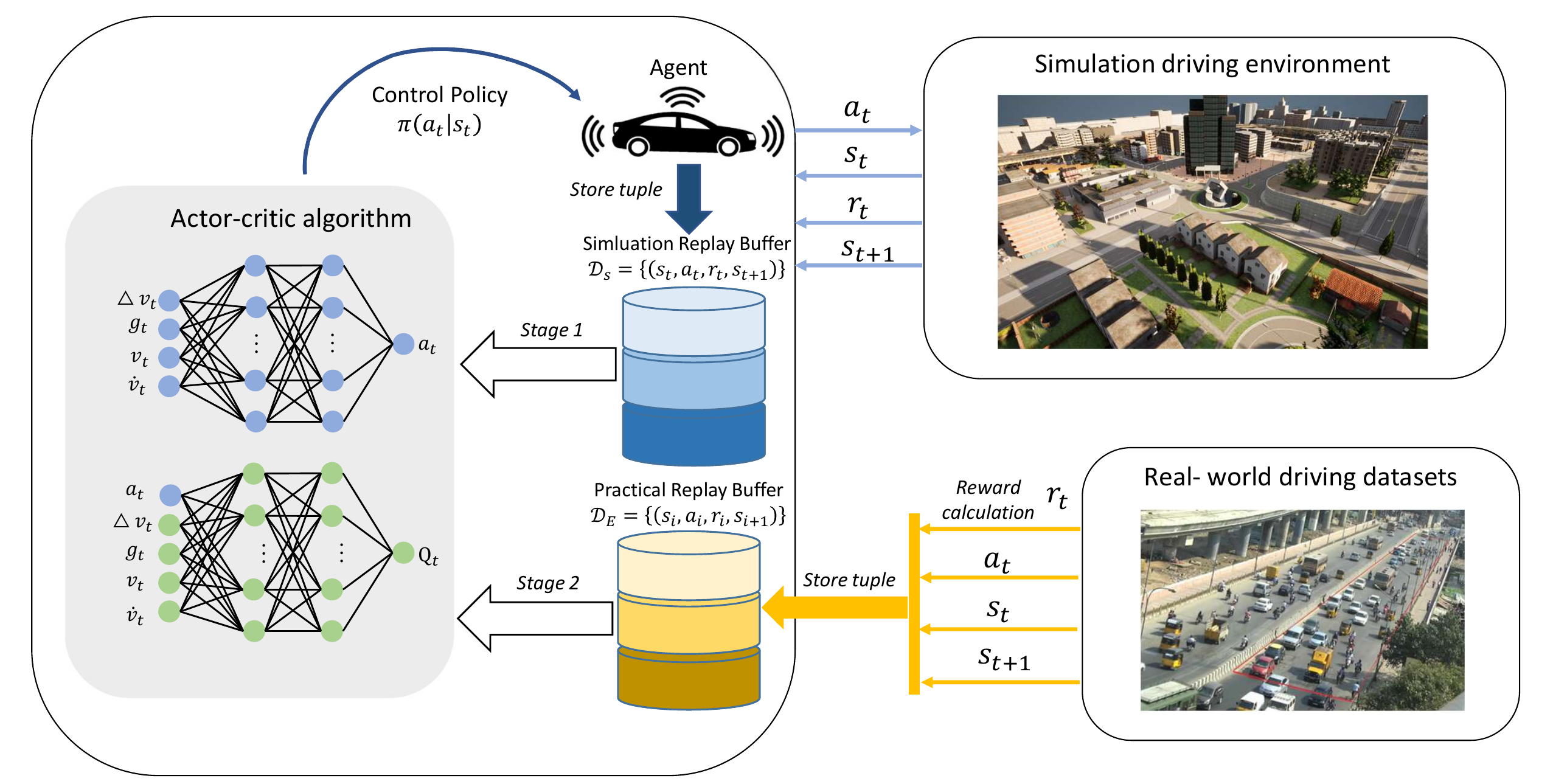}
    \caption{The conceptual framework of the two-stage DDPG agent. Stage 1: the agent updates networks with the simulation replay buffer. Stage 2: the agent also learns from the modified practical replay buffer based on real driving datasets.}
    \label{fig:The concept framework of this paper}
\end{figure}

All these limitations above motivated us to propose the two-stage DRL algorithm that tackles all of these issues. The two-stage DRL algorithm trains car-following agents that balance the policies learned by the agent itself through DRL in the simulator and the policies implied in real-world human demonstration datasets. Therefore we obtain an agent that is superior to pure DRL agents and drivers in the datasets and more suitable for real-world traffic. The overall system architecture is depicted in Figure \ref{fig:The concept framework of this paper}. First, we use DRL with experience replay \citep{lin1992self} to train an agent that can follow the leading vehicle and maintain the desired distance and speed. Second, we store the real driving datasets into the practical replay buffer and resume training the agent with both replay buffers. We itemized our contributions as follows:

\begin{itemize}
    \item [1.] We mix the real-world database with the self-generated data in the simulation, to encourage the DRL agent to learn expert-like behaviors by leveraging the mixed data (\citealp{hester2018deep}; \citealp{vecerik2017leveraging}). We prove that our two-stage DRL algorithm can obtain the expert-like behavior policy which is superior to pure DRL agents and real divers, and this will also tackle the performance drop due to dataset with limited size in common DRL or BC algorithms.
    
    \item [2.] Since DRL agents need to continuously interact with the environment to obtain rewards and optimize their policy, which means training a DRL agent in a real-world traffic environment is often impractical and expensive. Therefore, we propose a novel framework that uses \href{https://carla.org/}{CARLA (Car Learning to Act)}, an open-source simulator for autonomous driving research, and \href{https://www.ros.org/}{Robot operating system (ROS)} for co-simulation to train our DRL agent in a real-world like environment with real dynamics and can be applied seamlessly to the real world with greatly improved efficiency (\citealp{Dosovitskiy17}; \citealp{quigley2009ros}). 
    \item [3.] We compare different utilization of real-world driving datasets for DRL agents and tackle the \textit{extrapolation error} when learning from real data.
    
\end{itemize} 

To sum up, within this paper we propose the first car-following model that is trained via real datasets and DRL on the simulating environment CARLA and ROS with hyper-realistic dynamics.
The remainder of this paper is organized as follows. In Section \ref{sec:Related works}, the preliminary knowledge related to this work is reviewed. Then, we introduce our approach in detail in Section \ref{sec:Our approaches} and the experimental setups for the training and evaluation process are discussed in Section \ref{sec:Experimental setup}. Afterwards, we evaluate the trained agents with different driving scenarios and discuss the results in Section \ref{sec:Evaluation and discussion}. Section \ref{sec:Conclusion} summarizes this paper.

\section{Related works}
\label{sec:Related works}

\subsection{Reinforcement Learning}
\label{subsec:Reinforcement Learning}
In RL, an agent interacts with an environment under the objective to maximize the received reward. Consider it as a Markov Decision Process (MDP), an autonomous agent at time step $t$ observes a state $s_t$ of the environment and then interacts with the environment by executing an action $a_t$ according to the policy $\pi(a_t|s_t)$. Afterwards, the environment and the agent transition to a new state $s_{t+1}$ with the probability $P(s_{t+1}|s_t, a_t)$ and meanwhile a reward $r_{t+1}$ is provided to the agent as the feedback. The objective is to find an optimal policy $\pi^*$ that maximizes the expected discounted cumulative rewards $\mathbb{E}_{\pi^*}\left\{ \sum_{k=0}^{\infty} {\gamma}^k r_{t+k+1} \right \}$, where $ {\gamma} \in \left[0, 1\right]$ is the discount factor with lower values making the agent to prefer immediate to distant rewards. Furthermore, in RL we use a state-value function $V^\pi$($s_t$) = $\mathbb{E}_\pi \left\{ R_t| s_t \right\}$ to measure how good a certain state $s_t$ is, in terms of expected cumulative reward by following a certain policy $\pi$. Similarly, the action-value function $Q^\pi$($s_t, a_t$) = $\mathbb{E}_\pi \left\{ R_t| s_t, a_t \right\}$ is defined as the expected return of the agent, that starts from state $s_t$, executes an action $a_t$, and follows policy $\pi$ afterwards. Although RL can come up with some optimal policies after training, for the large state and action space in high dimensions, such a tabular format becomes computationally infeasible. With the establishment of deep learning, DRL regained interest with DNN being used to approximate the optimal Q values (\citealp{mnih2015human}; \citealp{silver2016mastering}).

\subsection{Deep deterministic policy gradient (DDPG)}

To tackle the high-dimensional, continuous action spaces, \citet{lillicrap2015continuous} developed a Deep deterministic policy gradient (DDPG) method that can learn policies in high-dimensional, continuous action spaces. DDPG is an actor-critic algorithm with two networks: actor network $\mu(s| \theta^\mu)$ with network parameter $\theta^\mu$ and critic network $Q(s, a | \theta^Q)$ with network parameter $\theta^Q$. The actor network will output an action $a_t$ based on the given states $s_t$, the critic network gets the action $a_t$ from the actor network, and the given states $s_t$, predict the goodness of the action $a_t$. DDPG uses the target networks method with the target actor network $\mu^\prime$ and target critic network $Q^\prime$. \citet{lillicrap2015continuous} found that for more stable learning, it is better to make the target networks slowly track the trained networks. The parameters of the target networks after each update of the trained network are updated using a sliding average for both the actor and the critic:
\begin{align}
      \theta^{\mu^\prime} &= \tau \theta^\mu + (1-\tau)\theta^{\mu^\prime},\\
     \theta^{Q^\prime} &= \tau \theta^Q + (1-\tau)\theta^{Q^\prime},
\end{align}
where $\tau \ll 1$ is the soft target update rate. \par

To balance between exploration and exploitation, an additive noise is usually added to the deterministic action to explore the action space:
\begin{equation}
    a_t=\mu(s_t| \theta^\mu) + \xi_t. 
\end{equation}

This additive noise $\xi_t$ could be the realization of the Gaussian distribution or the zero-reverting \citet{uhlenbeck1930theory} process suggested by \citet{lillicrap2015continuous}. The pseudo-code of DDPG is shown in Algorithm \ref{algorithm:DDPG}. In this work, as the action and state space for the car-following task are continuous, we use DDPG as our DRL algorithm. Since the main contribution is the proposal of the fusion of RL and small real datasets with simulation environment CARLA, we do not make any comparison with other DRL algorithms.



\begin{algorithm}
\label{algo:ddpg}
\SetKwData{Left}{left}\SetKwData{This}{this}\SetKwData{Up}{up}
  \SetKwFunction{Union}{Union}\SetKwFunction{FindCompress}{FindCompress}
  \SetKwInOut{Input}{input}\SetKwInOut{Output}{output}
\caption{DDPG algorithm}\label{algorithm:DDPG}
Initialize actor network $\mu$ and critic network $Q$ with random weights $\theta^\mu$ and $\theta^Q$ \par
Initialize the target networks $\mu^\prime$ and $ Q^\prime$ with weights  $\theta^{\mu^\prime} \gets \theta^\mu$, $\theta^{Q^\prime} \gets \theta^Q$\par
Initialize replay buffer ${D}$ \par
\For{$ episode \in \left[ 1, M \right] $}{
    Initialize a random process noise $\xi$ for action exploration \par
    Receive initial observation state $s_1$ \par
    \For{$t\in \left[ 1, T \right]$}{\label{forins}
      Select action $a_t=\mu(s_t| \theta^\mu) + \xi_t $ according to the current policy and exploration noise \par
      Execute action $a_t$ and observe reward $r_t$ and observe new state $s_{t+1}$\par
      Store transition $(s_t, a_t, r_t, s_{t+1})$ in ${D}$ \par
      Sample a random minibatch of $N$ transitions $(s_i, a_i, r_i,s_{i+1})$ from ${D}$ \par
      Set $y_i = r_i + \gamma Q^\prime \left\{s_{i+1}, \mu^\prime(s_{i+1}|\theta^{\mu^\prime})|\theta^{Q^\prime}\right\}$ \par
      Update critic by minimizing the loss: $L = \frac{1}{N} \sum_{i} \left\{y_i - Q(s_i, a_i| \theta^Q)\right\}^2$ \par
      Update the actor policy using the sampled policy gradient:
      $\nabla_{\theta^{\mu}} J \approx {\frac{1}{N} \sum_{i} \nabla_a Q(s, a | \theta^Q)|}_{s = s_i, a = \mu(s_i)} {\nabla_{\theta^{\mu}} \mu(s| \theta^{\mu})}|_{s_i} $ \par
      Update the target networks:\par
      $\theta^{\mu^\prime} = \tau \theta^\mu + (1-\tau)\theta^{\mu^\prime}$\par
      $\theta^{Q^\prime} = \tau \theta^Q + (1-\tau)\theta^{Q^\prime}$ 
      
    }
  }
\end{algorithm}

\subsection{Experience replay}
If the networks only learned the highly correlated consecutive samples from the experience as they occurred sequentially in the environment, this would therefore lead to inefficient learning. One possible approach to tackle this problem involves the use of \emph{experience replay} proposed by \cite{lin1992self}. Following it we store the experiences of an agent at each time step in a dataset called the \textit{replay memory} ${D}_t$. Let the experience of an agent at time $t$ can be denoted as:
\begin{equation}
    d_t = (s_t, a_t, r_t, s_{t+1}),
\end{equation}
containing the state of the environment $s_t$, the action $a_t$ from agent at given state, the reward $r_t$ the agent received as a result of the previous state-action pair $(s_t, a_t)$, and the next state of the environment $s_{t+1}$. With this replay buffer, we randomly sample some tuples that break the temporal correlations by mixing less recent experiences for the updates, and rare experiences will be used for more than just a single update. To utilize real driving datasets in DRL, \emph{experience replay} mechanism is essential, 
since we need to store the dataset in the \textit{replay memory} and let the agent learn the driving policy implicit in the datasets.

\subsection{Extrapolation error}
\label{Extrapolation error}

Extrapolation error is the distributional shift between the dataset and true state-action visitation of the current policy in off-policy value learning \citep{fujimoto2019off}.
Since in RL, the accuracy of Q-function depends on the estimate of the Q-value, when the current policy $\pi(a|s)$ differs substantially from the policy $\pi_{\beta}(a|s)$ in the replay buffer, which means the target policy selects an out-of-distribution (OOD) action $a^\prime$ at the state $s^\prime$, such that $(s^\prime, a^\prime)$ is unlikely, or not contained in the replay buffer. The estimate of Q-value may be arbitrarily bad without
sufficient data near the current $(s^\prime, a^\prime)$ and these estimation errors will accumulate over each iteration, resulting in arbitrarily poor performance \citep{levine2020offline}. \par

As stated by \citet{fujimoto2019off}, the off-policy RL algorithms are ineffective when learning \textit{truly off-policy} without any online interaction, not only the distributional shift of trained policy but also the distributional shift of different initialization will affect the performance of off-policy learning. One way to overcome this problem is to add active data collection and compensate the estimated error (\citealp{vecerik2017leveraging}; \citealp{hester2018deep}; \citealp{dadashi2021offline}). For this reason, we combine the real-world driving dataset and the online interactive experience in the replay buffer to tackle \textit{extrapolation error} and improve the performance of our DRL agent.

\subsection{CARLA and ROS}
\label{CARLA}
Since training a DRL agent in the real world is impossible and expensive, we usually use a simulated environment instead. There are numerous well-known autonomous driving simulators used in academic research such as CARLA, CarSim, LGSVL, Gazebo, etc. In this work, we use CARLA and ROS as our training platforms. CARLA is an open-source simulator grounded on Unreal Engine with hyper-realistic physics that uses the OpenDRIVE standard to define roads and urban settings (\citealp{Dosovitskiy17}; \citealp{dupuis2010opendrive}). It is based on a scalable client-server architecture, with the server being responsible for simulation tasks, e.g. sensor rendering, computation of physics, etc. The client is usually defined by the user, which via API can easily specify the sensor setup of the vehicle. Afterwards, a controlled amount of camera images and vehicle-related information will be provided for further usage. Because of the high fidelity of CARLA, it is often used in the field of autonomous driving (\citealp{ravi2018real}; \citealp{dworak2019performance}; \citealp{gomez2020train}; \citealp{tran2019robust}; \citealp{niranjan2021deep}). Furthermore, an important reason why we choose CARLA as our training environment is that it provides the CARLA-ROS bridge that enables two-way communication between Robot Operating System (ROS) \citep{quigley2009ros} and CARLA. \par

Connection to ROS is vital for our work since ROS is an open-source robotics middleware suite, which enables the communication between various sensors on a single robot and multiple robots without affecting the independence of each part. As our vehicles are equipped with several sensors to perceive the surroundings, exposing nearby vehicles, communication middleware is required between vehicles as well as between sensors and controllers within one vehicle. Since automated vehicles are complex systems with a high degree of interdependencies between their components, ROS is suitable for us to carry out autonomous driving tasks such as car-following and makes our agent vehicle access the information of the leading vehicle such as position and velocity in a leader-follower pair \citep{hellmund2016robot}. Furthermore, ROS is a robot system that is well developed and used for autonomous robots and vehicles. The use of ROS in the simulation would help us to transfer the trained car-following agent to the real vehicle seamlessly in our future work.

\section{Our approaches}
\label{sec:Our approaches}
\subsection{System overview}

As shown in Figure \ref{fig:The concept framework of this paper}, we train and evaluate our control policies directly using the CARLA simulator. The agent interacts with the simulation environment to obtain rewards, observe the states, and store them in the replay buffer. Therefore we update the DNN with the small batches sampled from the replay buffer. Here, we assume that the states observed by the agent are already known, including the position of the leader and follower, velocity, acceleration, and other information. Since this only involves simple sensor information processing issues which are out of the scope of the current research. After training the DRL agent, we store the processed real driving dataset into another replay buffer called the \emph{practical replay buffer} and continue training the agent with both replay buffers. 


\subsection{DDPG agent for longitudinal control }
\label{subsec:Deep Deterministic Policy Gradient(DDPG)}
In this paper, we use DDPG as our working DRL algorithm to obtain the longitudinal control policy, with the setup introduced in this section.

\subsubsection{Action and state space}
\label{sec:Action and state space}

In the case of the end-to-end DRL algorithm, the action space is the throttle and brake. Although this can potentially solve the internal control problem of the vehicle, it is accompanied by a decrease in generalization. Different vehicles have different dynamic characteristics, therefore keeping our algorithm useful for fixed vehicles and cannot be applied to other types. For this reason, the action space $A$ of our DDPG agent uses continuous variable acceleration $\dot{v}_t$. Moreover, the real human driving dataset of the platoon driving experiments of \cite{punzo2005nonstationary} does not contain direct information about throttles and brakes, but the speed and acceleration. If we want to use real driving data, it is also more conducive to using acceleration as action space.\par
To follow the leader, the agent needs to be able to communicate with it, so as the state space $S$, we use such features as the velocity of follower $v_t$, the acceleration of follower $\dot{v}_t$, the leader's velocity $v_{t,l}$, and the bumper-to-bumper gap $g_t$. As suggested by \citet{kim1999normalization}, we normalize the data generally for speeding up the learning process and faster convergence. The state $s_t$ at time step $t$ is thus defined as 
\begin{eqnarray}
s_t = \begin{pmatrix}
   \frac{v_t}{v_{des}}\\
   \frac{\dot{v}_t - \dot{v}_{min}}{\dot{v}_{max} - \dot{v}_{min}}\\ 
    \frac{v_{t,l} - v_t}{v_{des}} \\
    \frac{g_t}{g_{max}} \end{pmatrix}, 
\end{eqnarray}
where $v_{des}$ denotes the desired speed, $\dot{v}_{max}$ and $\dot{v}_{min}$ define the feasible range of accelerations, $g_{max}$ indicates the maximum space gap for normalization of the input states.

\subsubsection{Reward function}
\label{sec:Reward function}
As the most critical part of the RL algorithm, the formulation of the reward function is highly related to the performance of the agent. We choose the reward function proposed by \citet{hart2021formulation}, with which, the trained agent performs greatly in the car-following task. The reward function focuses on safety factor and also makes a balance between the comfort for the driver and the driving efficiency. It can be divided into three weighted sub-rewards:

\begin{equation}
r_{t} = w_{safe}r_{t,\;safe} + w_{gap} r_{t,\;gap} + w_{jerk} r_{t,\;jerk}.
\label{eq:reward}
\end{equation}
\\
The first term $r_{t,\;safe}$ compares the kinematically needed deceleration with the comfortable deceleration $b_{comf}$ and focuses on the response of driver to safety-critical situations: 

\begin{equation}
r_{t,\;safe} =-\tanh \left(\frac{b_{kin}-b_{comf}}{-\dot{v}_{min}}\right) \mathbb{I} {\left\{b_{kin} > b_{comf}\right\}},
\label{eq:safe}
\end{equation}
where $b_{kin} = \frac{v_t - v_{t,l}}{g_t}  \mathbb{I}{\left\{v_t > v_{t, l}\right\}}$, and $\mathbb{I}{\left\{ \cdot \right\}}$ is the indicator function.

The second term $r_{t, \;gap}$ devises for driving efficiency, to prevent the follower agent from keeping a long distance with the leader to avoid the occurrence of dangerous situations:
\begin{equation}
r_{t,\;gap} =
\begin{cases} 
\frac{\varphi \left\{ (g_t - g_{opt})/g_{var}\right\}}{\varphi(0)} , & \text{if}\                                g_{t} < g^{*}, \\
\frac{\varphi \left\{ (g_t - g_{opt})/g_{var}\right\}}{\varphi(0)}\left(1 - \frac{g_t - g^*}{g_{lim} - g^*}\right) , & \text{otherwise},
\end{cases}
\label{eq:r_t_gap}
\end{equation}    
where $g_{opt} = v_t T + g_{min}$, $g_{var} = 0.5g_{opt}$, $g_{lim} = v_t T_{lim} + 2g_{min}$, and $\varphi \left\{ x \right\}$ describing the density function of the standard normal distribution. 

The third term $r_{t,\;jerk}$ addresses the control of jerk for a comfortable driving:

\begin{equation}
r_{t,\;jerk} = - \left(\frac{1}{j_{comf}} \frac{\mathrm{d}\dot{v}_{t}}{\mathrm{d}t}\right)^2.
\end{equation}

As the weights measure the trade-off among different sub-rewards, we conducted a series of experiments, tested different combinations of weights, and selected the set of weights that has the best performance and at most interpretable. We choose the weights for safety factor $w_{safe} = 1.0$, efficiency factor $w_{gap} = 0.5$ and the comfortable factor weights $w_{jerk} = 0.004$ with the rest of hyperparameters in Table \ref{tab:Hyperparameters used for the reward function setup}. As seen from the weights, safety, being the most important, has the highest weight, while jerk, being just a comfort parameter the lowest.

\begin{table*}[htb]
    \centering
    \begin{tabular}{cll}
    \Xhline{1pt}
    Symbol  & Description   & Value \\
    \hline
    $\dot{v}_{min}$  & Minimum acceleration & \SI{-9}{m/s^2}\\
		$\dot{v}_{max}$ & Maximum acceleration & \SI{5}{m/s^2}\\
		${v}_{des}$  & Desired velocity & \SI{20}{m/s}\\
		$g_{max}$  & Maximum distance between leader and follower & 200 m\\
		$b_{comf}$  & Comfortable deceleration   & \SI{2}{m/s^2}\\
		$T$  & Desired time gap to the leading vehicle   & 1.5 s\\
		$g_{min}$  & Desired minimum space gap   & 2 m\\
		$T_{lim}$  & Upper time gap limit for zero reward   & 15 s\\
		\Xhline{1pt}
    \end{tabular}
    \caption{Hyperparameters used for the input states and reward function setup.}
    \label{tab:Hyperparameters used for the reward function setup}
\end{table*}

\subsubsection{Neural network structure}
\label{Neural network structure}
\begin{figure}
    \centering
    \includegraphics[width=3.5in]{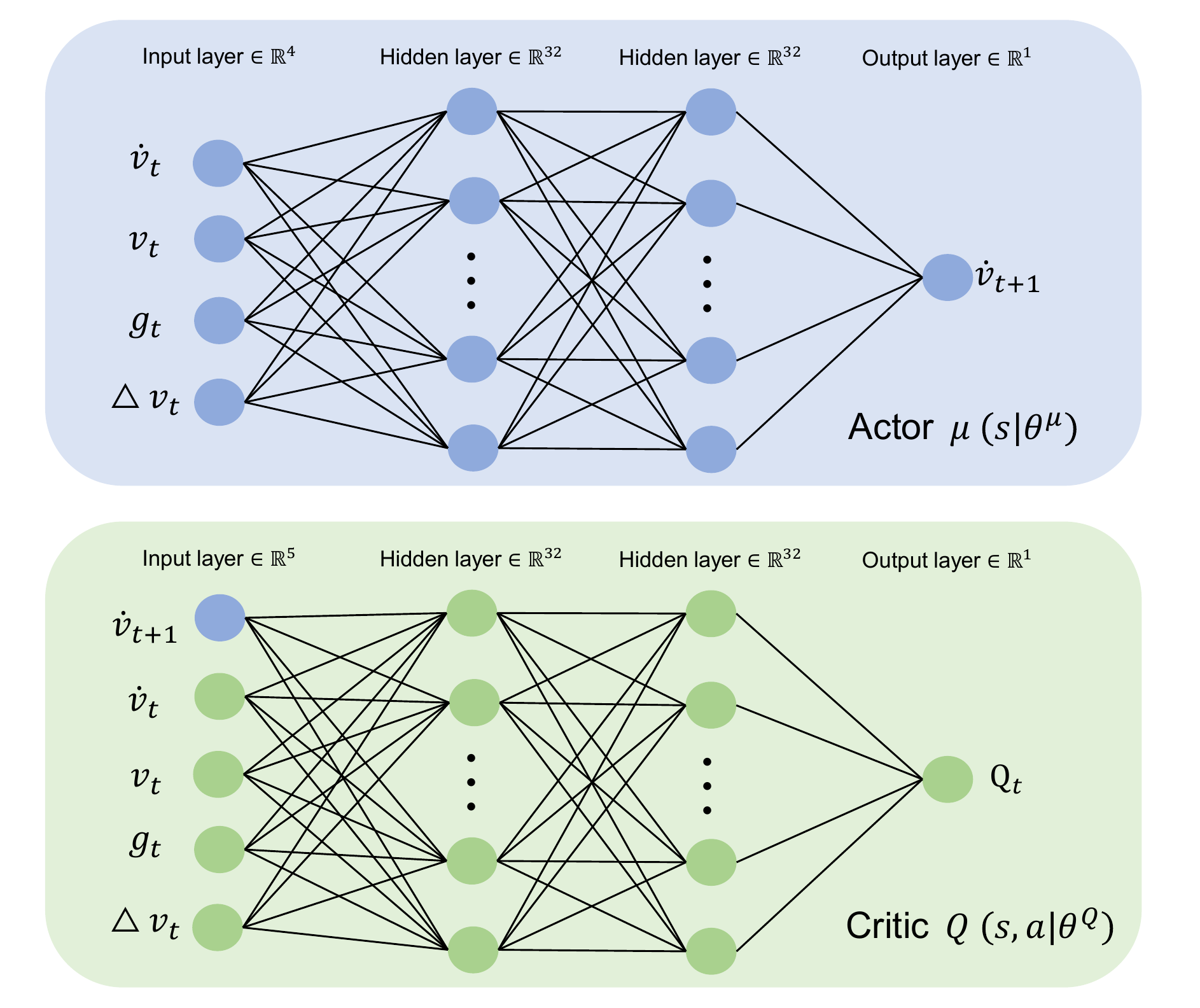}
    \caption{Neural network architecture of Actor and Critic in DDPG, each of them has two hidden layers with 32 neurons in each layer.}
    \label{fig:DDPG neural network architecture}
\end{figure}

The structure of DDPG neural networks used in this study is depicted graphically in  Figure \ref{fig:DDPG neural network architecture}. Both actor and critic are using fully connected neural networks with two hidden layers, both hidden layers containing 32 neurons. In two hidden layers, we use ReLU activation function. For the output layer in critic network, we use linear activation function, and for the output layer in actor network, we use $Tanh$ as activation function to scale the output to [-1, 1] for better performance, afterwards, this bounded output will be mapped to the acceleration range. 

\subsection{Lateral control policy}

In the car-following task,  we were usually only concerned about longitudinal control, but because of the high fidelity of physic parameters in the CARLA simulator, during the training, there were frictions between the tires and the road, which cause very small heading drift error, usually $0.5^{\circ}$ for 1000 training steps. To correct this heading drift error and eliminate effects on longitudinal control, we apply Stanley controller proposed by \citet{thrun2006stanley} as the lateral control policy (steering angle) to keep the vehicle in the middle of the road. The considered controller is one type of geometrical path-tracking controller, which helped Stanford to win the DARPA challenge in 2006. 
Moreover, since it considers both the lateral and the heading error, it shows a very good performance. In a Stanley controller, the steering angle $\phi(t)$ of the vehicle is given by:
    
\begin{equation}
\phi_t = \theta_{p}(t) + \tan^{-1} \left\{\frac{k_v \;d_{f}(t)}{v(t)}\right\}.
\notag
\end{equation}
where $\theta_{p}(t)$ compensates the angular error ${\theta_p}$ and the second term compensates the front lateral distance error ${d_f}$ measured from the center of the front axle to the nearest point on the path. Furthermore, $\frac{v}{k_v}$ is a headway distance and $k_v$ is a normalizing constant.

\subsection{Real-world human driving datasets}

We use the real-world car-following trajectory from \citet{punzo2005nonstationary} as our real-world human demonstration, which we later on call "Napoli datasets". It is created and processed after a platoon driving experiment, in which several vehicles follow each other on urban roads and highways. The Napoli datasets were collected on two days along the same route on two urban roads and one rural road. It consists of five sub-dataset with an overall amount of 24 min driving data. The first urban road is a 2 km long straight road with congested traffic (i.e., stop-and-go traffic conditions). It has four intersections and an estimated capacity of 900 vehicles per hour (veh/h). The second urban road is also approximately 2 km long but with an estimated capacity of 1200 veh/h.
The rural road is a two-lane 3 km long highway with an estimated capacity of 1500 veh/h. The traffic flow during collecting datasets was approximately 400 veh/h. With a non-stationary Kalman filter \citep{kalman1960new} applied to position data, high-quality car-following data was provided, including the velocity of each vehicle and bumper-to-bumper distance between leader-follower pair with the frequency of 10 Hz. \par
\begin{figure}
    \centering
    \includegraphics[width=0.8\textwidth]{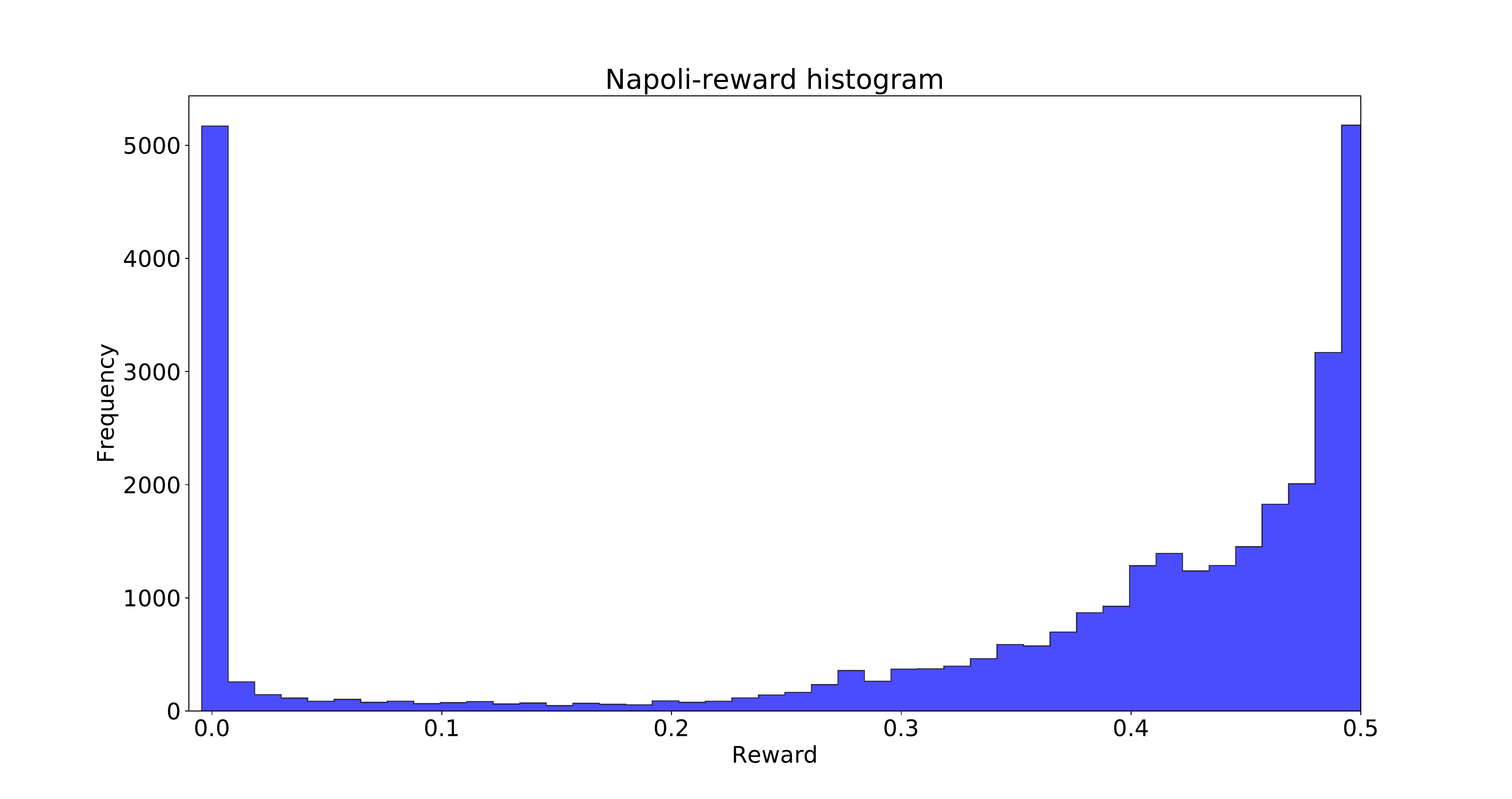}
    \caption{Reward distribution in Napoli datasets according to the reward function in Section \ref{sec:Reward function}.}
    \label{fig:Napoli_reward}
\end{figure}

  
To store the driving datasets into replay buffer, we need to construct lists of tuples $d_i$ = ($s_i$, $a_i$, $r_i$, $s_{i+1}$). Since the states and actions at each time step are already known from the datasets, only rewards are missing. To compute the reward for each action the human driver made, we use the same reward function (\ref{eq:reward}) as for the DDPG agent. 
The histogram of the resulted rewards for the human drivers is shown in Figure \ref{fig:Napoli_reward}. As the maximum reward $r_t$ agent can obtain is 0.5, we see that under this reward function setting, half of the actions performed by the human driver can be considered as good ($r_i \geq 0.4$). This means that we can expect that the agent will learn some good behaviors from human drivers. Nevertheless, we still observe a large fraction of actions with zero reward, which indicates that these actions executed by human drivers result in a higher time gap above the upper time gap limit in (\ref{eq:r_t_gap}) and Table \ref{tab:Hyperparameters used for the reward function setup}. We separate the Napoli dataset into two parts, 95$\%$ of the data for training, and 5$\%$ of the data for evaluation. For the second stage of training, we utilize training data and let the agent improve its driving policy by leverage between real driving data and online experience from interaction with the simulator. Furthermore, the real driving actions with lower rewards work as negative incentives which can exclude some undesirable behaviors. Moreover, the Next Generation Simulation (NGSIM) trajectory dataset is used as another real-world driving dataset for evaluation in Section \ref{sec: ecaluation ngsim}.\par

\section{Experimental setup}
\label{sec:Experimental setup}
\subsection{Simulation environment}
\begin{figure}
        \centering 
        \subfloat[The common driving environment]{\includegraphics[width=3.5in]{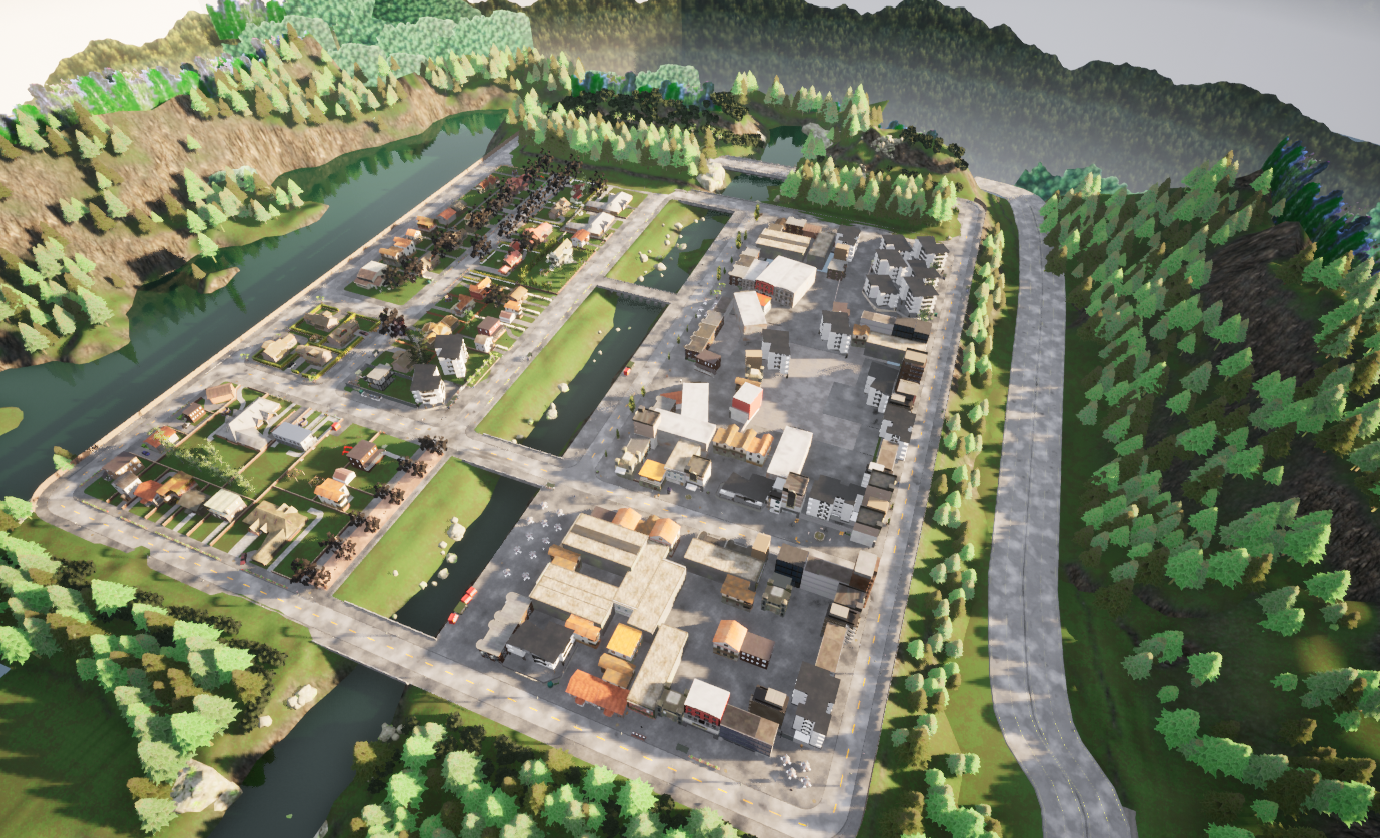}}
            \hfil 
         \subfloat[The OpenDRIVE standalone mode]{
            \includegraphics[width=3.5in]{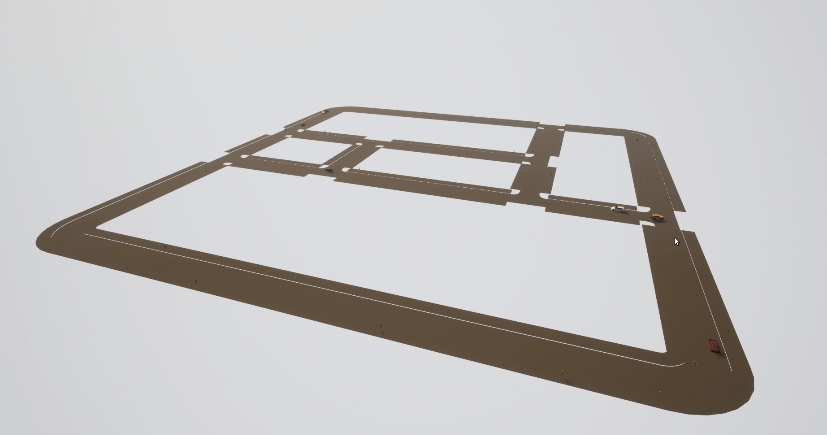}}
            \hfil
        \caption{Two different driving simulation modes of CARLA: (a) the standard driving mode in CARLA with full rendering; (b) the OpenDRIVE standalone mode with minimalistic rendering for better simulation efficiency.} 
        \label{fig:carla simulator}
\end{figure}

As discussed in Section \ref{CARLA}, we use CARLA being our simulation environment. Since CARLA is grounded on Unreal engine, the static objects like infrastructure, buildings, or vegetation, and dynamic objects such as vehicles, cyclists, or pedestrians are made by 3D models. For the car-following task, we just need leading and following vehicles running on the road. To save computing resources and speed up the algorithms, we choose the OpenDRIVE standalone mode of CARLA for simulation. The standalone mode runs the full simulation by only using the OpenDRIVE file which describes every detail on the road. Other additional geometries or assets such as buildings or vegetation will not be created, as shown in Figure \ref{fig:carla simulator}. The simulator will take the OpenDRIVE \textit{.xodr} file and procedurally create temporal 3D meshes which describe the road definition in a minimalistic manner. Moreover, to prevent vehicles from falling off the road, visible walls are created at the boundaries of the road, to act as a safety measure.


\subsection{CARLA-ROS co-simulation}
\label{CARLA-ROS co-simulation}
Although the hyper-realistic simulated environment of CARLA provides API to modify all aspects related to the simulation, to process the sensor information, and enable interoperability with control and perception modules, we should rely on ROS. In this paper, we use CARLA-ROS co-simulation to implement and evaluate our approach. For the car-following task, we need to ensure the communication between leader and follower, which ROS has strength at \citep{hellmund2016robot}. Furthermore, ROS provides GUI tools for diagnostics and monitoring during the development and run-time of the system. This allows us to have an immediate overview of the status of the system. \par

The transmission of information between two vehicles as well as the control node in ROS is realized via a transport system with \textit{publish} / \textit{subscribe} semantics. The control node can obtain required information from each vehicle, and publish control information to the agent, for instance, throttle and brake.\par

\begin{figure}
    \centering
    \includegraphics[width=\textwidth]{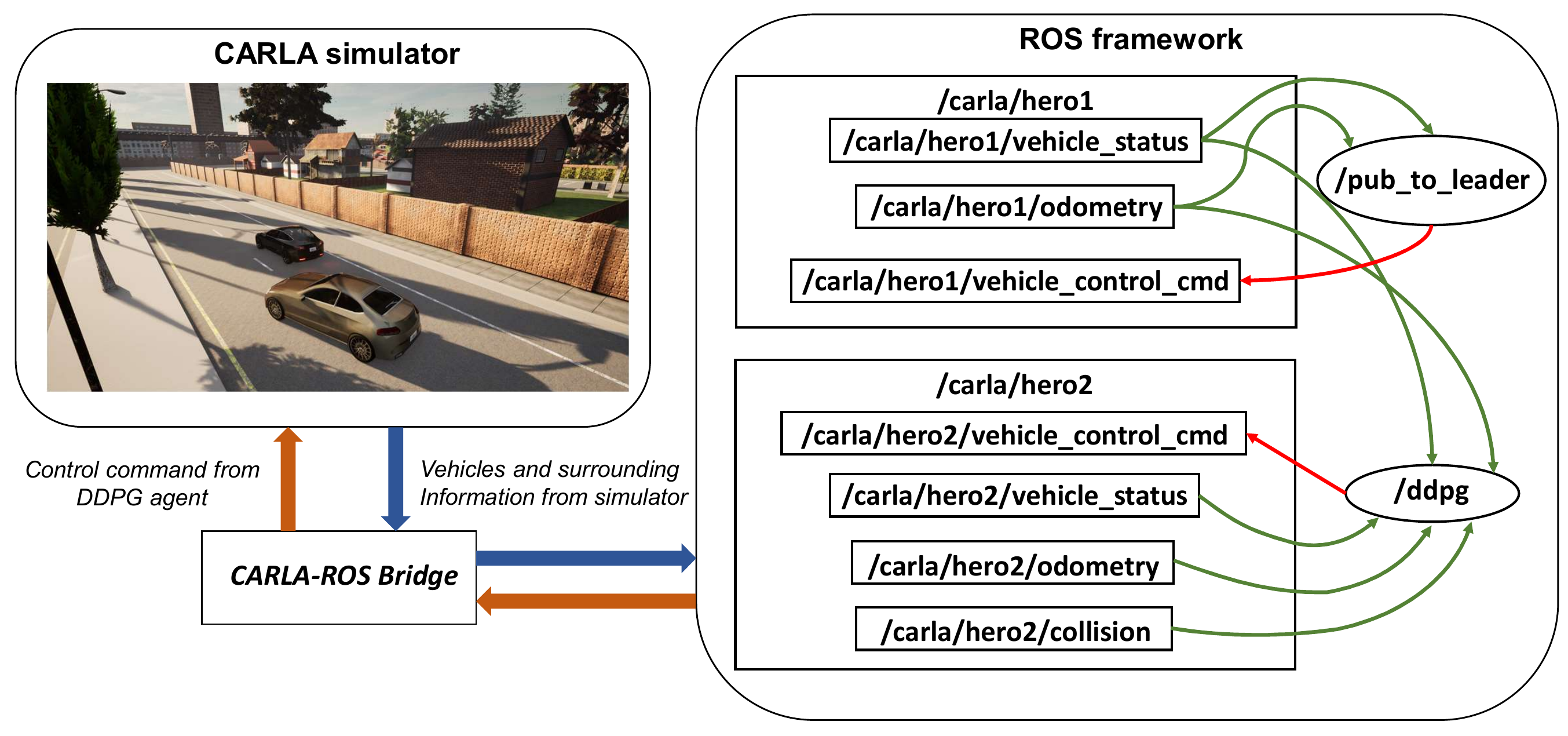}
    \caption{The structure of our proposed CARLA-ROS framework, the CARLA-ROS bridge transport the information of vehicles and surrounding from CARLA into ROS, afterward, transfer the control command from DDPG agent in ROS to the vehicle in CARLA.} 
    \label{fig:ROS node graph for the DDPG}
\end{figure}

Here, \textit{hero1} indicates the leader, and \textit{hero2} means the follower in a leader-follower pair. Four different topics provide different information about the vehicles in CARLA:

\begin{itemize}
    \item [1.] \textit{/carla/hero/vehicle\_status}:  provides the current speed, acceleration, orientation of the vehicle, as well as the control values reported by CARLA.
    \item [2.] \textit{/carla/hero/odometry}: provides the current position of the vehicle in the specified coordinate frame which usually is Cartesian coordinates (UTM).
    \item [3.] \textit{/carla/hero/vehicle\_control\_cmd}: messages sent to CARLA apply a control signal to the vehicle, which includes throttle: [0.0, 1.0], brake: [0.0, 1.0], steering: [-1.0, 1.0].
    \item [4.] \textit{/carla/hero/collision}: retrieves collision data detected by the collision sensor. It is used to reset the environment when a collision between two vehicles occurs.
\end{itemize}

As mentioned in Section \ref{subsec:Deep Deterministic Policy Gradient(DDPG)} and shown in Figure \ref{fig:ROS node graph for the DDPG}, neural networks that were updated and learned will output a control signal, such as acceleration of the agent, and will be transported back to CARLA to control the agent. However, in CARLA, there is no controller to directly use acceleration or speed to control vehicles which are our actions in DDPG. This can be performed only indirectly via throttle and brake. For this reason, we use the reverse data method to obtain a controller that controls vehicles via acceleration and speed, as discussed in details in the next section.

\subsection{Controller for CARLA}
\label{Controller for CARLA}
For CARLA simulation, \textit{/carla/hero/vehicle\_control\_cmd} topic is used to control the vehicle via throttle and brake. But as discussed above, we should use acceleration as the output of our agent to ensure generalization. To overcome this problem, there are two feasible solutions: First, as vehicle models in CARLA are wrappers for the PhysX Vehicle by NVIDIA GameWorks, we can map the throttle and brake to velocity and acceleration with lists of equations according to the vehicle dynamic models. Nevertheless, in many cases, we can not use this method, when the dynamic models of the object are unknown, for instance, with real vehicles connected via ROS. Second, we can use the reverse data method to train a simple \emph{control neural network} via supervised fashion that links acceleration and velocity to throttle and brake. In this paper to allow for generalization, we go for the second approach. \par
At first, we run four hours of driving test with automatic control mode in CARLA. In this case, the vehicle is controlled automatically by CARLA with throttle and brake. During the driving, we collect datasets that include the velocity and acceleration of the vehicle at each time step, as well as the throttle and brake at the same time step that result in the corresponding acceleration and velocity. Second, we reorganize the collected dataset with the input states: 
    \begin{eqnarray}
    x_t = \begin{pmatrix}
   v_{t+1}\\
   v_{t}\\
   a_t \end{pmatrix},
   \notag
    \end{eqnarray}
where $v_{t+1}$ is the target velocity of next time step, $v_{t}$ is the current velocity, $a_t$ is the current acceleration. The output states of the neural network are: 
    \begin{eqnarray}
    y_t = \begin{pmatrix}
   g_{t}\\
   b_{t}
   \end{pmatrix},
   \notag
    \end{eqnarray}
where $g_{t}$ and $b_{t}$ are throttle and brake at time step $t$ respectively. This procedure can be performed with any other simulation environment.\par
\begin{figure}
    \centering
    \includegraphics[width=3in]{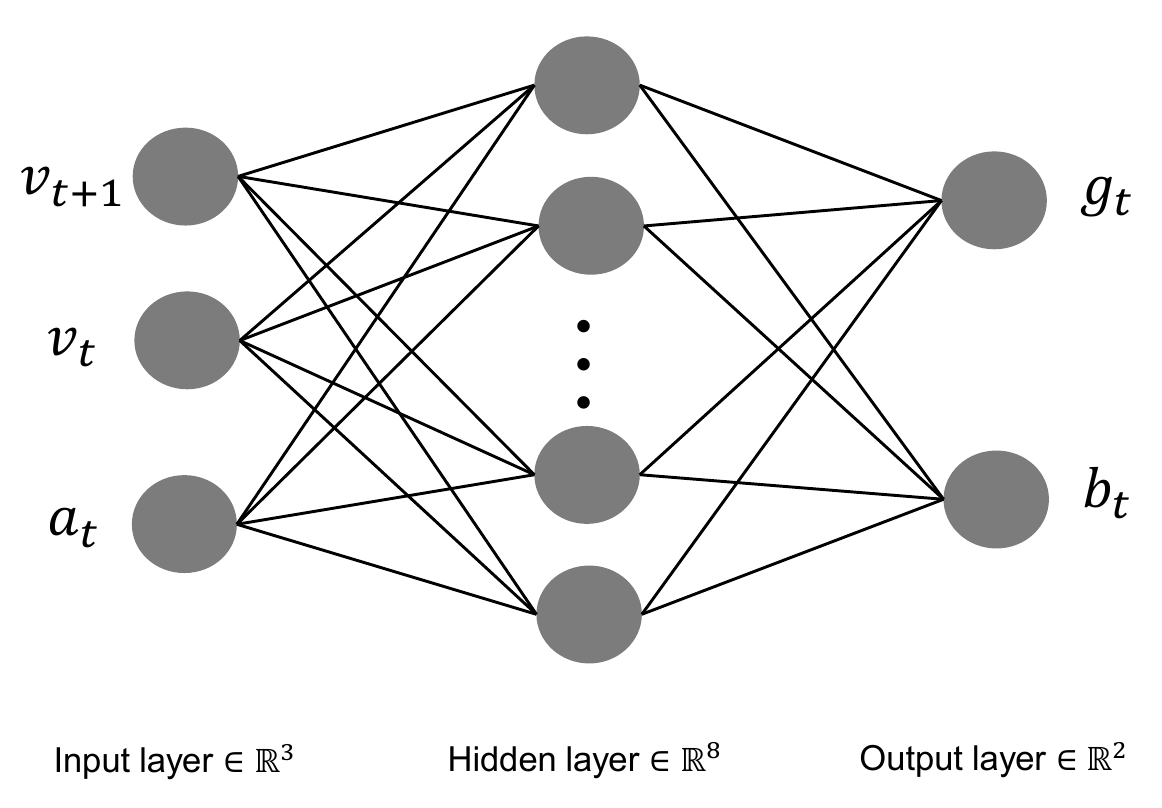}
    \caption{Control neural network with inputs of $v_{t+1}$, $v_{t}$, $a_t$, and output $g_t$ and $b_t$.}
    \label{fig:Control neural network}
\end{figure}

We use a simple fully connected neural network as shown in Figure \ref{fig:Control neural network}, and train it with the new reorganized dataset. Afterwards, if the proposed DDPG algorithm is planned to be applied to different models of vehicles, only this control neural network should be retrained. As an example, if we have a robot car in the real world, but the actual physic models are unknown. We can first collect the controlling data similarly as in CARLA, and then train the control network with the input of acceleration. Afterwards, this extra control network can be connected to our pretrained DRL policy network and resume training for some small episodes to achieve sufficient performance. Compared with the end-to-end DRL, this method greatly improves generalization. \par

\subsection{Training process}
\label{sec:Training process}

\begin{figure}
    \centering
    \includegraphics[width=\textwidth]{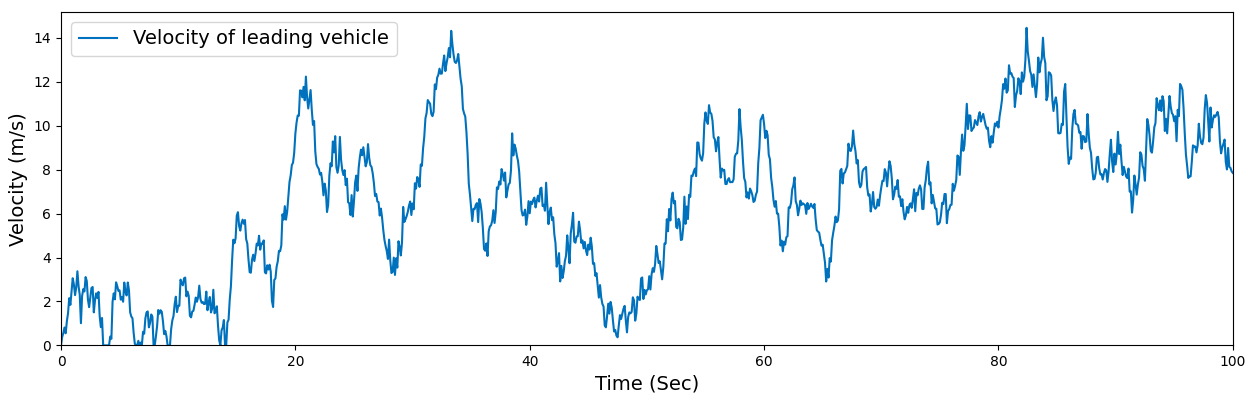}
    \caption{An example leading vehicle velocity trajectory based Ornstein-Uhlenbeck process for the RL training process.}
    \label{fig:leader velocity}
\end{figure}

To train car-following agents, leader-follower pairs are required. Thus, the Ornstein-Uhlenbeck process is used to generate random leader trajectories, an example leader velocity trajectory is illustrated in Figure \ref{fig:leader velocity}. Before each episode of the RL training process, such random velocity trajectories will be generated, afterwards, the ROS node \textit{pub\_to\_leader} receives the velocity commands from the generated trajectories and controls the leading vehicle to run with the desired velocity in CARLA. The agent aims to follow the leading vehicle with the obtainment of maximum rewards according to the reward function defined in Section \ref{sec:Reward function}. Furthermore, the initial speeds of the leader and the follower are set to \SI{0}{m/s}, and the initial bumper-to-bumper gap is set randomly in the range $\left[\SI{0}{m}, \SI{100}{m} \right]$. \par

To balance exploration and exploitation, we add exploration noise following another zero-reverting Ornstein-Uhlenbeck process to the output of the DDPG policy network as suggested by \citet{lillicrap2015continuous}, and we use clipping to ensure that the processed output does not exceed the boundary. The structure of actor and critic neural networks is discussed in Section \ref{Neural network structure}. Before training our DDPG agent, we deliberately choose a list of hyperparameters for the learning as shown in Table \ref{tab: the simulation}.

\begin{table*}
	\centering
	\begin{threeparttable}
	\begin{tabular}{cll}
		\Xhline{1pt}
		Symbol  & Description   & Value \\
		\hline
		$l_r$  & Learning rate & 0.001\\
		$\gamma$ & Reward discount factor & 0.95\\
		$N_D$  & Size of the replay buffer & 2000 \\
		${B}$  & Training batch size & 32\\
		$\tau$  & Soft target update rate   & 0.001\\
		$\theta_1$  & Parameter in Ornstein–Uhlenbeck process   & \SI{0.15}{s^{-1}}\\
		$\sigma_1$  & Parameter in Ornstein–Uhlenbeck process   & \SI{0.20}{s^{-0.5}}\\
		\Xhline{1pt}
	\end{tabular}
	\begin{tablenotes}
        \scriptsize
        \item[1.] $\theta_1$ and $\sigma_1$ were determined according to the suggestion in \citep{lillicrap2015continuous}.   
      \end{tablenotes}
	\caption{Hyperparameters used in the simulation.}
	\label{tab: the simulation}
	\end{threeparttable}
\end{table*}

We compare several approaches with different utilization rates for the real human driving dataset, to cover the whole spectrum between DRL and BC, and also include the IDM model \citep{treiber2000congested}:

\begin{enumerate}
    \item [$\bullet$] BC: train a neural network same as the actor network in DDPG with real-world human driving datasets by supervised learning.
    
    \item [$\bullet$] IDM: the IDM parameters shown in Table \ref{tab: idm} are calibrated according to the evaluation datasets.
    
    \item [$\bullet$] Fully off-policy DDPG: fill the replay buffer only with real-world human driving datasets and train the DDPG agent from scratch.

    \item [$\bullet$] Pure DDPG: only trained by interacting with the simulator, without any real-world human driving datasets.
    
    \item [$\bullet$] Two-stage DDPG: after training a pure DDPG, ratio $r$ of the replay memory is sampled from \emph{practical replay buffer} which consists of real-world human driving datasets, and other $(1-r)$ from \emph{simulation replay buffer} with self-generated experience by interacting with the environment. Afterwards, training is resumed. In this work, we choose the ratio $r$ from 0.1 to 1.0 with 0.1 step to cover the whole spectrum.
    
\end{enumerate}

\begin{table*}
	\centering
	\begin{threeparttable}
	\begin{tabular}{cll}
		\Xhline{1pt}
		Symbol  & Description   & Value \\
		\hline
		$v_{des}$  & Desired velocity & \SI{20}{m/s}\\
		$T$ & Safe time gap & \SI{1}{s}\\
		$a$  & Maximum acceleration & \SI{2}{m/s^2} \\
		$b_{comf}$  & Comfortable deceleration & \SI{2}{m/s^2}\\
		$g_{min}$  & Minimum gap   & \SI{2.5}{m}\\
		\Xhline{1pt}
	\end{tabular}
	\caption{Hyperparameters used for IDM model.}
	\label{tab: idm}
	\end{threeparttable}
\end{table*}

\begin{figure}
    \centering
    \includegraphics[width=4in]{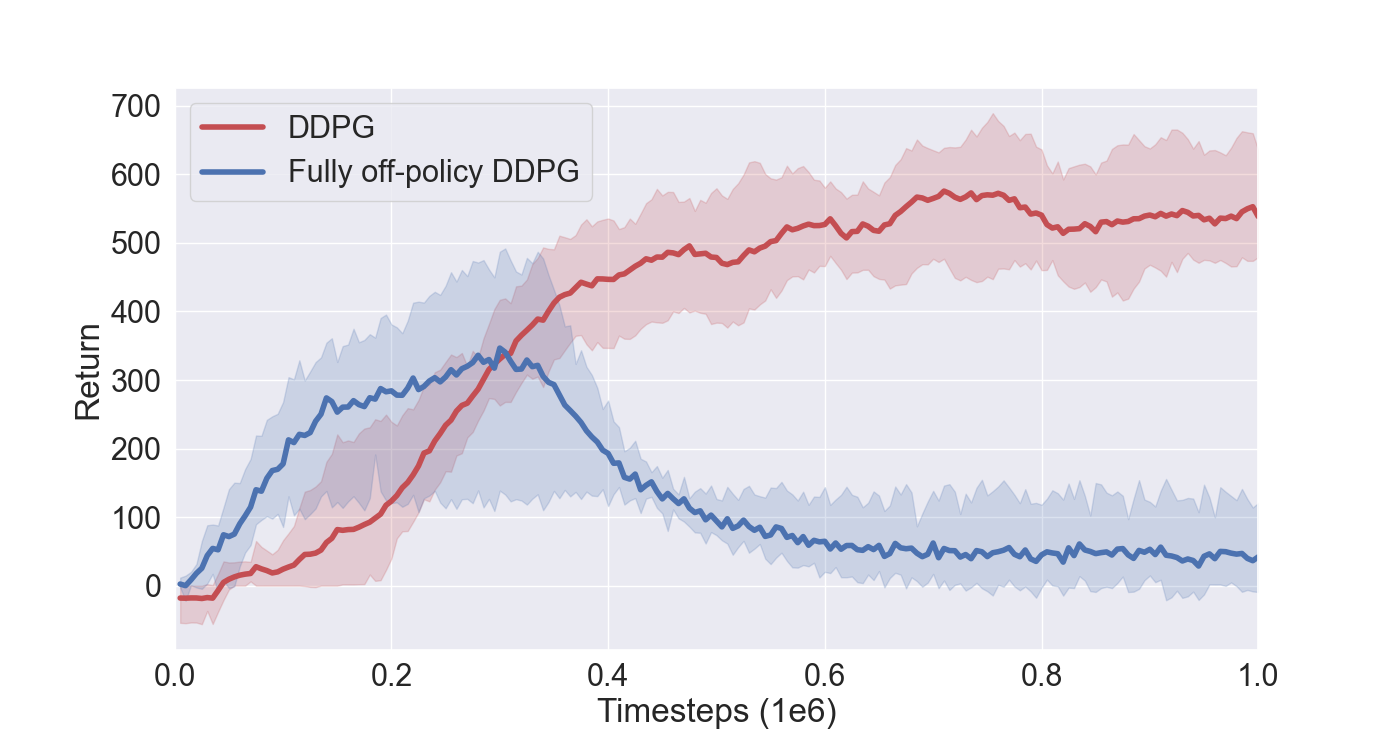}
    \caption{Rewards obtained by the pure DDPG agent and fully off-policy DDPG agent during the training process.}
    \label{fig:reward plot}
\end{figure}

The training process runs on an NVIDIA GeForce RTX 3080 GPU and takes roughly two hours to finish 10000 steps of interaction with the simulated environment. The simulation interval is 0.1 seconds, which means the DRL agent makes decisions every 0.1 seconds corresponding to 10 HZ of real data observation. Our DDPG agents and fully off-policy DDPG are trained for one million timesteps multiple times with different initial seeds, with the reward obtained over time shown in Figure \ref{fig:reward plot}. We see that the DDPG agent is converged during the training, with even more training steps, the performance will not be improved. However, the fully off-policy DDPG agent fails to learn from the real data. This result suggests that differences in the state distribution under the initial policies cause extrapolation error, which can drastically offset the performance of the fully off-policy agent. Furthermore, the BC agent is trained for 20 epochs with the real-world human driving dataset. \par

\section{Evaluation and discussion}
\label{sec:Evaluation and discussion}
We apply the proposed methods and train an autonomous driving agent in the designed driving scenarios. Afterwards, methods are evaluated on the other set of scenarios.

\subsection{Evaluation with a leader trajectory from real-world driving dataset}

\begin{figure}
        \newcommand{\w}{0.45}
        \centering 
        \subfloat[Speed comparison of different followers]{\includegraphics[width=\textwidth]{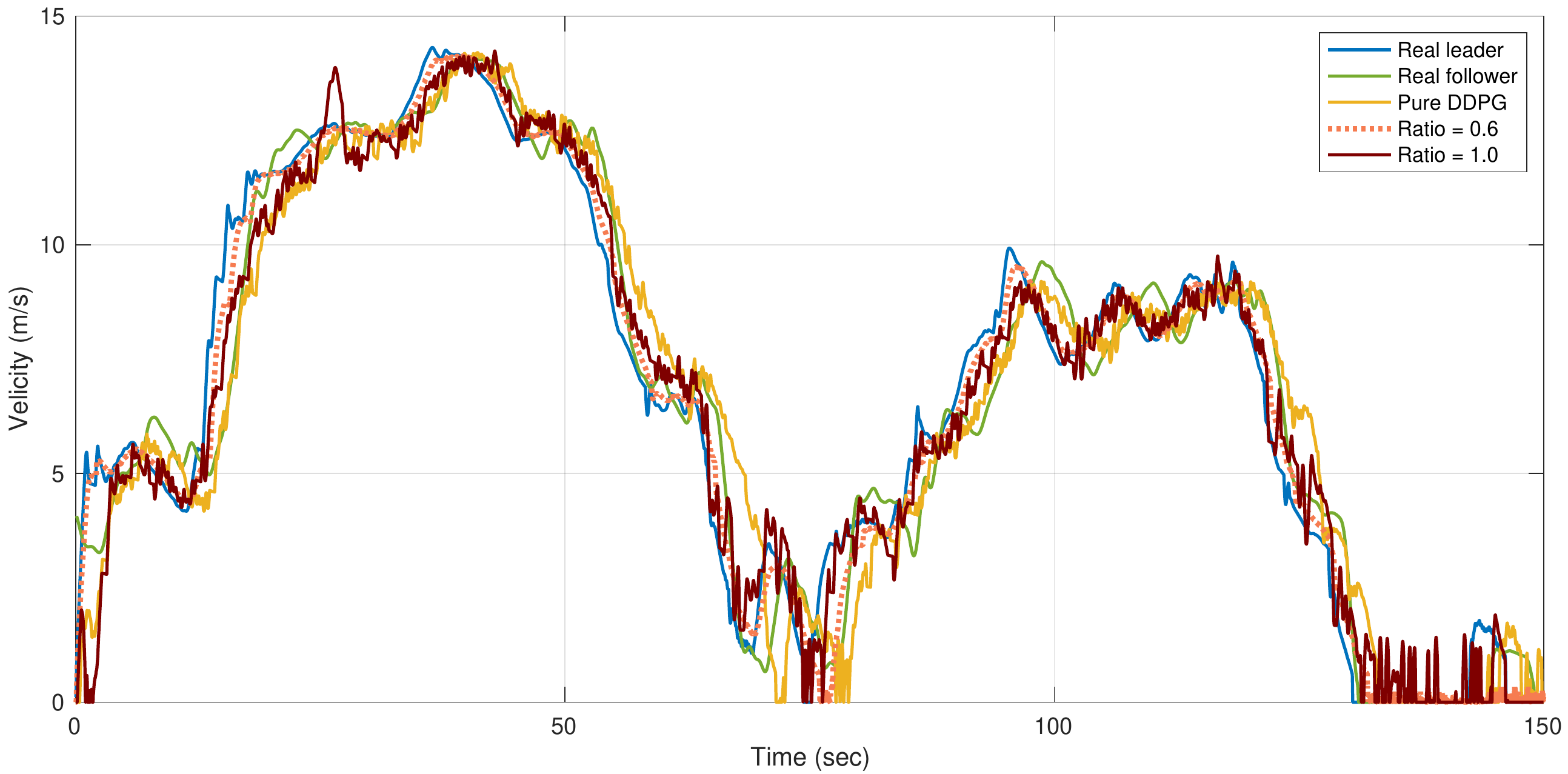}}
            \hfil 
         \subfloat[Bumper-to-bumper gap comparison of different followers]{
            \includegraphics[width=\textwidth]{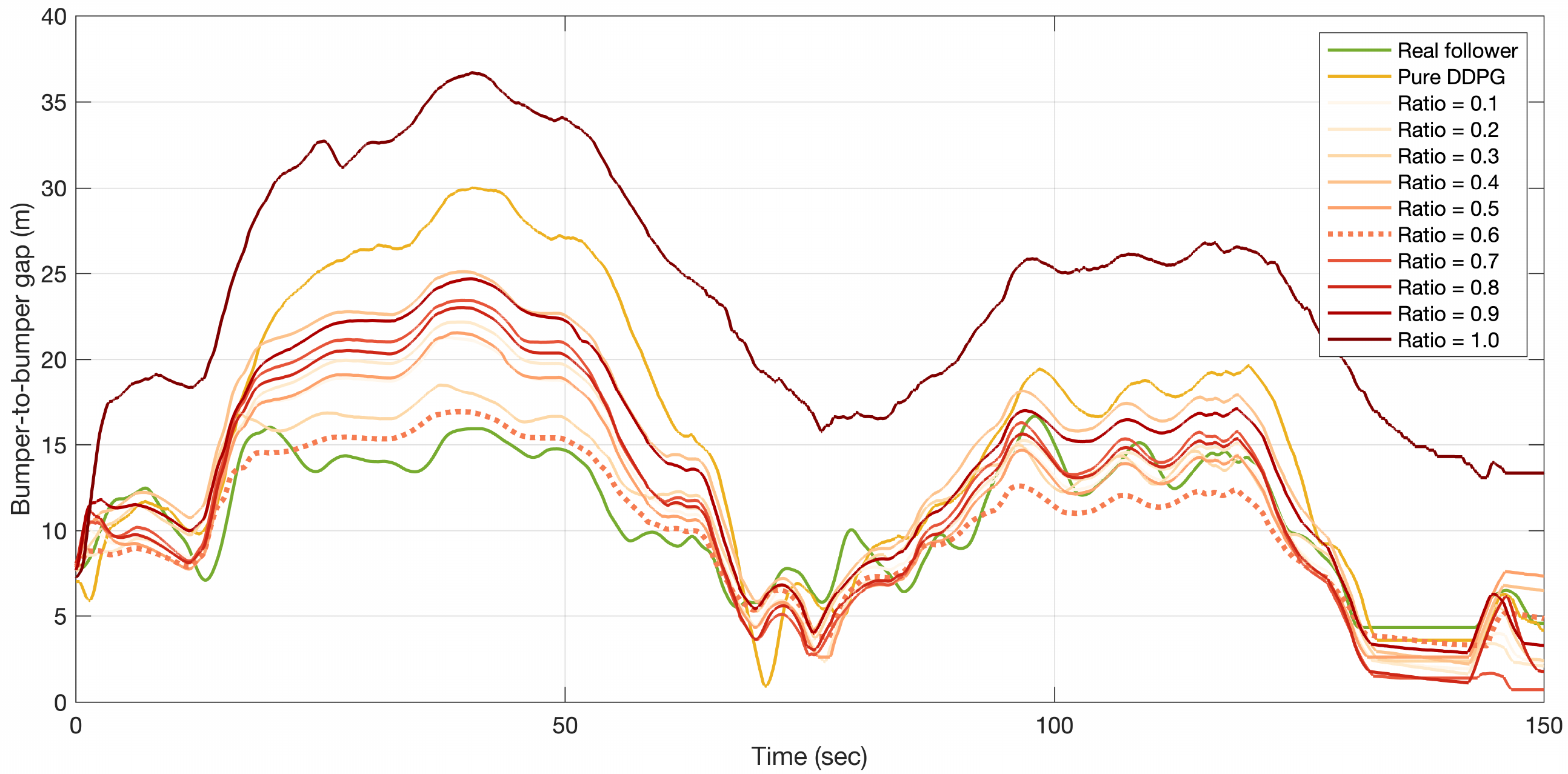}}
            \hfil
        \caption{The evaluation results by following a real leader trajectory (blue) from Napoli datasets for two-stage DDPG followers with different utilization ratios of the real dataset. For better visualization, the velocity trajectories in (a) just show the $Ratio = 0.6$ and $Ratio = 1.0$ compared with pure DDPG and real follower, with the rest of the velocity trajectories shown in \ref{sec:appendix a}.}
        \label{fig:different ratio}
\end{figure}

First, after the training process discussed in Section \ref{sec:Training process}, we get several two-stage DDPG car-following agents. To verify that the trained two-stage DDPG agents learned human driving behaviors from real data, we compare the trained two-stage DDPG followers with pure DDPG follower and the real follower by following the evaluation leader trajectory from the Napoli dataset which has a car-following period of 150s. The results are depicted in Figure \ref{fig:different ratio} and \ref{sec:appendix a}, where the velocity trajectory and the bumper-to-bumper distance between each follower and their leader are used. We see that the trained two-stage agents can follow the trajectory of the leading vehicle, whether it is in acceleration, deceleration, or standstill phases. The DDPG follower with 100\% real data i.e. $Ratio = 1.0$, performs worse than the pure DDPG follower and keeps too much distance to the leading vehicle, moreover, the velocity profile of this follower is unstable with too much unnecessary acceleration and deceleration. The DDPG agent with the ratio of 1.0 real data starts from the pure DDPG, after training a pure DDPG, fill the replay buffer only with real-world human driving datasets, and resume training. Even with pretty good initial performance, the discrepancy between the trained policy and the policy in the datasets can offset the performance. On contrary, other two-stage DDPG followers also start from the pure DDPG agent but learn from real-world human demonstrations and interactive experiences. Even with different proportions of the real dataset i.e. from $ Ratio \in \left\{0.1, \ldots, 0.9\right\}$, with the help of the interactive experience from the training process, the estimation error from OOD will be corrected. Therefore, the two-stage DDPG improves its driving behavior from pure DDPG follower by learning from the real dataset. Nonetheless, different proportions of the real dataset in the replay buffer still cause different performances. The two-stage DDPG with $Ratio = 0.6$ has the best performance compare with other followers. In the high-speed driving phase, for driving efficiency, the vehicle maintains an appropriate distance while ensuring safety, and in the standstill phase keep a more safe distance to the leading vehicle. The two-stage DDPG agents modified its driving policy with the real driving dataset and tends to drive much closer to the leader compared with the pure DDPG agent. Noticing that during the standstill phases such as (T$\in$[75s, 80s]) and (T$\in$[130s,140s]), when the leader is waiting for the traffic light in the signalized intersection, the DDPG with human experience agent still keeps small velocity to approach the leader with safe distance, and wait for the traffic light turning green. For further evaluation, we use the two-stage DDPG agent with $Ratio = 0.6$ as our working two-stage DDPG agent.

\subsubsection{Compare with other approaches}
\label{sec:compare with other approaches}
\begin{figure}
        \newcommand{\w}{0.45}
        \centering 
        \subfloat[Speed comparison of different followers]{\includegraphics[width=\textwidth]{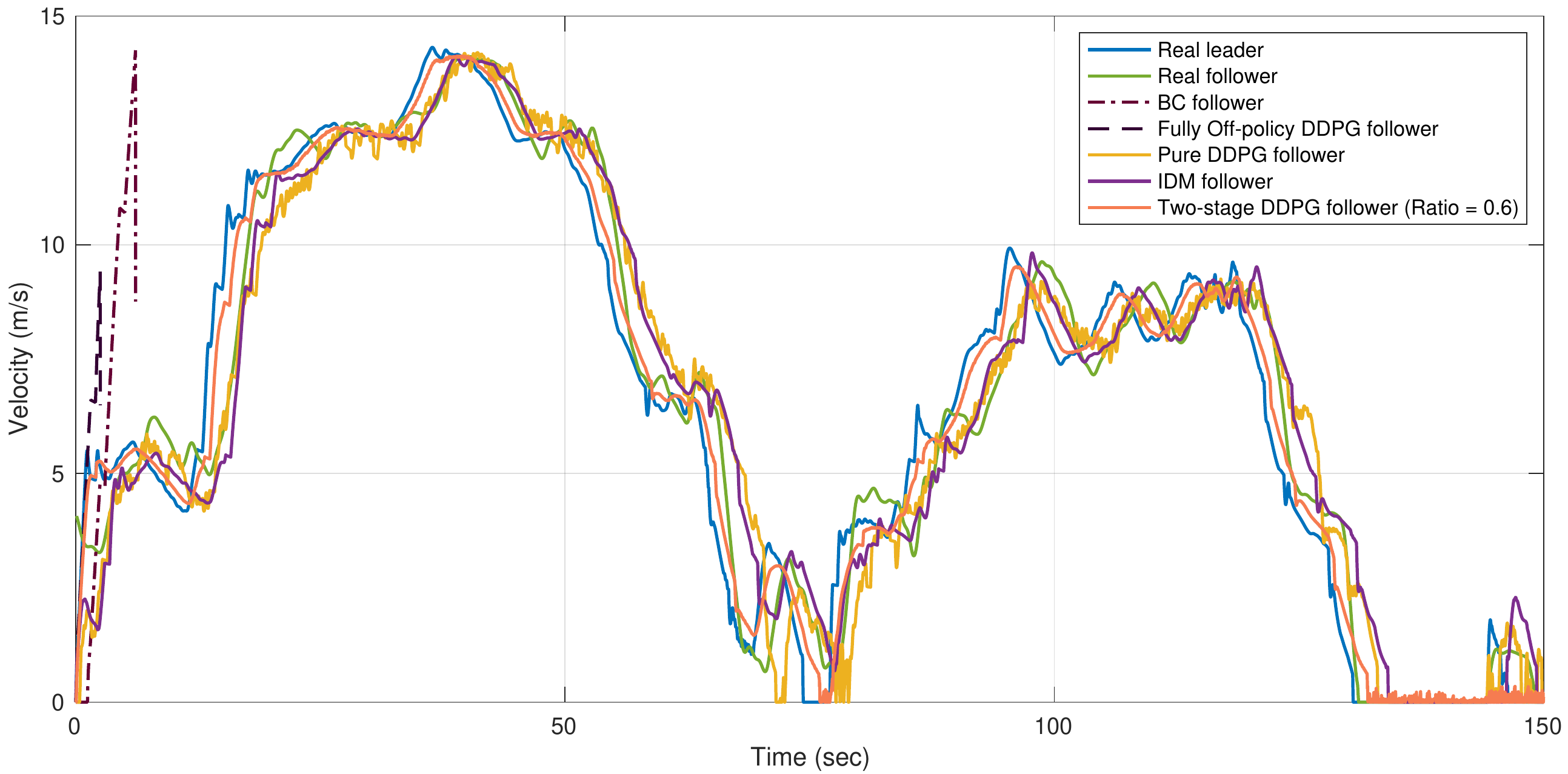}}
            \hfil 
         \subfloat[Bumper-to-bumper gap comparison of different followers]{
            \includegraphics[width=\textwidth]{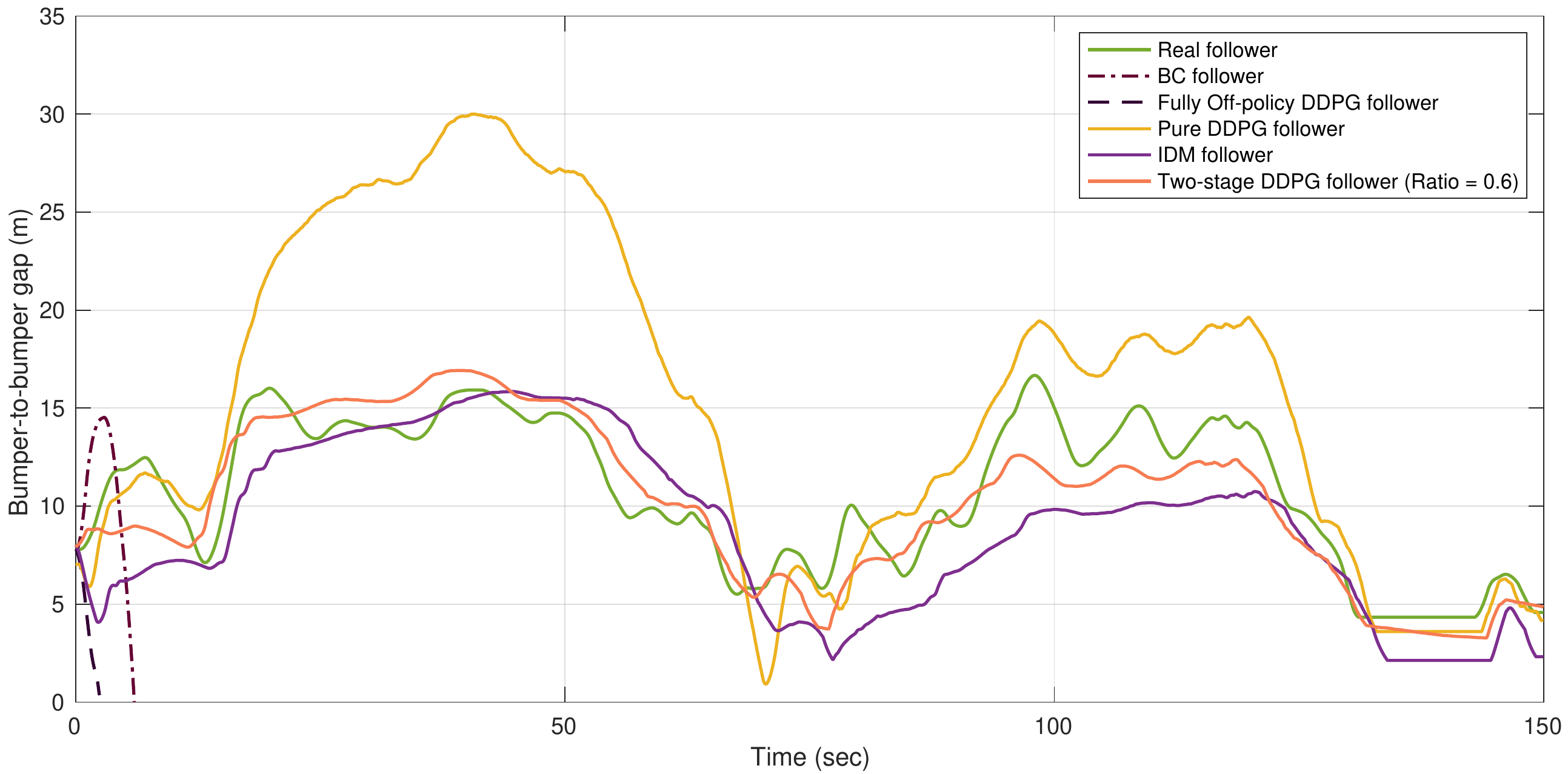}}
            \hfil
        \caption{The evaluation results by following a real leader trajectory (blue) from Napoli datasets for different followers.} 
        \label{fig:real_leader_with_2ddpg}
\end{figure}

Since we have compared different two-stage DDPG agents with different utilization ratios of the real dataset, a horizontal comparison with other approaches is also necessary. Here we compare the Two-stage DDPG agent ($Ratio = 0.6$) with BC agent, fully off-policy agent, IDM follower, and real follower.
As illustrated in Figure \ref{fig:real_leader_with_2ddpg}, we see that the BC agent and fully off-policy agent fail at the beginning of the evaluation. Fully off-policy DDPG agent suffer from extrapolation error that we introduced in Section \ref{Extrapolation error}, where the agent learns only with the real-world datasets from scratch, which means the action distribution generated by the initial policy has enormous difference to the action distribution of the real-world datasets, leading to a no-learning result. \par

The agent trained through the reward function shows a more passive driving style during the high-speed driving phase (T$\in$[15s, 65s] and T$\in$[90s, 125s]) and tends to be safe, therefore keeping a larger distance from the leader. However, in the low-speed or standstill phase (T$\in$[130s, 140s]), the IDM model accustom to keep a short distance to the leader, which is a potential safety risk for not only the passenger of the ego vehicle but also for the leading vehicle driver. These immature driving behaviors are not suitable for real-world traffic. The two-stage DDPG agent modified its driving policy and tends to drive much closer to the leader compared to the pure DDPG agent during the high-speed phase, during the low-speed phase or standstill phase, the two-stage DDPG agent tries to stand farther to the leader than IDM follower. 
Moreover, from Figure \ref{fig:real_leader_with_2ddpg} we see that the two-stage DDPG follower has a shorter reaction time than other car-following agents. 

\subsubsection{Time-to-collision (TTC) analysis}

\begin{figure}
        \newcommand{\w}{0.45}
        \centering 
        \subfloat[Distribution of TTC under 100s]{\includegraphics[width=4.5in]{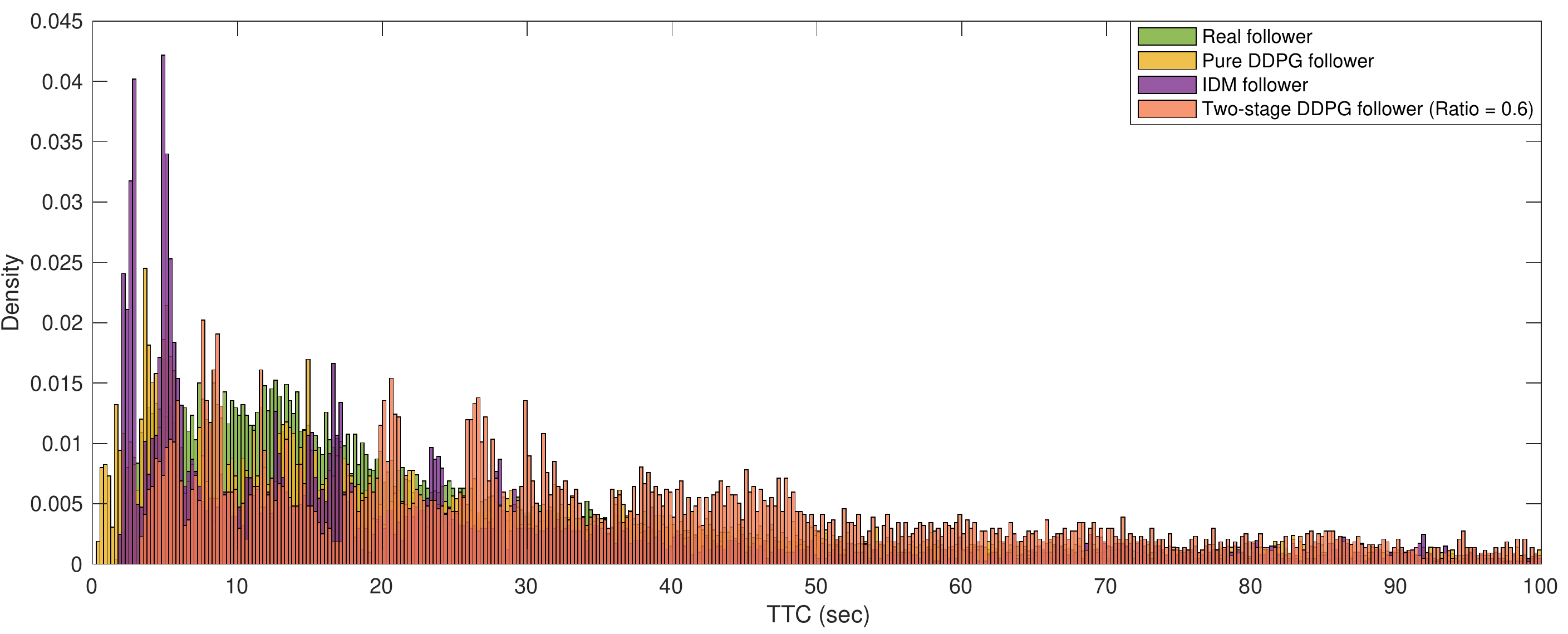}}
            \hfil 
         \subfloat[Comparison for three agents with TTC less than 5s]{
            \includegraphics[width=4.5in]{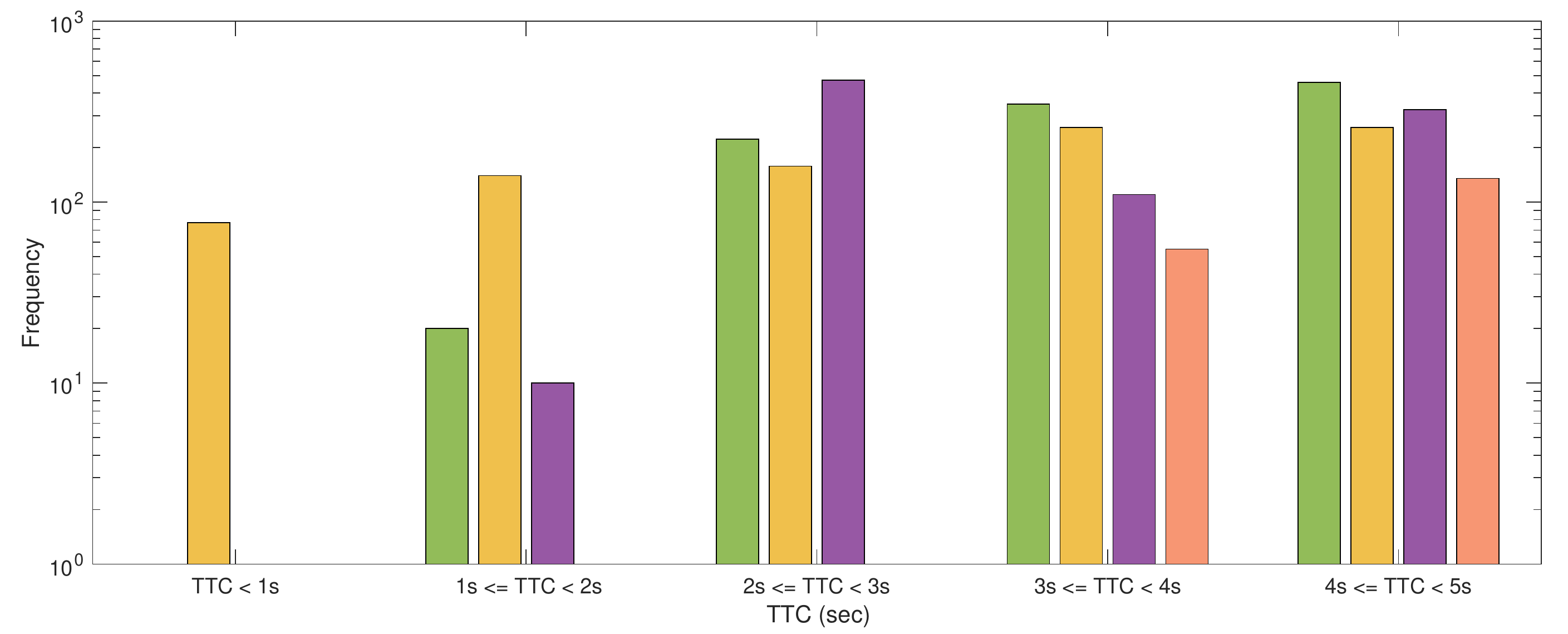}}
            \hfil
        \caption{Distribution of Time-to-collision (TTC) for real follower in dataset (green), pure DDPG agent (yellow), IDM (purple),  two-stage DDPG agent (orange).} 
        \label{fig:TTC_real_leader}
\end{figure}

TTC is a widely used proximal safety indicator initially introduced by \citet{Hayward1972NEARMISSDT}. Figure \ref{fig:TTC_real_leader} shows the TTC below 100 seconds of every time step for real follower from dataset, pure DDPG agent, two-stage DDPG agent, and the IDM model. Table \ref{tab:ttc} shows the TTC distribution under 10s for different followers. Since the fully off-policy DDPG agent and BC agent either has worse performance than pure DDPG or fails in the evaluation, these agents are excluded in TTC analysis.\par
We consider TTC under 2s as the safety-critical situation \citep{minderhoud2001extended}. As illustrated in Figure \ref{fig:TTC_real_leader}, the pure DDPG follower, IDM model as well as the real follower suffer from TTC lower than 2s, which indicate dangerous situations, but our two-stage DDPG follower has a higher lower bound than the other three agents. The lower TTC bound of the pure DDPG agent, real follower and IDM model is 0.48s, 1.73s and 1.98s respectively, while the one for the two-stage DDPG agent is 3.37s. Moreover, the TTC distribution of the two-stage DDPG is mainly concentrated around 7s with smaller variance considering TTC under 10s. The two-stage DDPG follower not only improves its driving policy with the real dataset but also learns to refrain from risky driving behavior implied in the real dataset. Thus, we see that the two-stage agent maintains a larger TTC than the pure DDPG agent, real follower, and IDM model while driving, which indicates better safety.

\begin{table*}
    \centering
    \begin{tabular}{lcccc}
    \Xhline{1pt}
     & \multicolumn{4}{c}{TTC (Sec)}\\
    \cline{2-5}
    Agents  & Minimum & Mean & Median & Std. dev\\
    \hline
    Real driver & 1.73 & 6.27 & 6.33 & 2.21\\
    IDM   & 1.98 & 4.90 & 4.88 & 2.02\\
    Pure DDPG  & 0.48 & 4.87 & 4.73 & 2.42\\
	Two-stage DDPG  & 3.37 & 6.94 & 7.44 & 1.77\\	
	\Xhline{1pt}
    \end{tabular}
    \caption{Computational results of different agents with TTC under 10s.}
    \label{tab:ttc}
\end{table*}


\subsection{Evaluation with NGSIM dataset}
\label{sec: ecaluation ngsim}
The NGSIM trajectory dataset is a well-known and de facto standard database, providing longitudinal and lateral positional information for about 3366 vehicle trajectories in certain spatiotemporal regions. In this work, we use NGSIM I-80 trajectory dataset as another evaluation database to compare the performance of the baseline algorithms and our two-stage DDPG follower. For the car-following task, since the drivers on the road are affected by the complex traffic environment, we need re-extract leader-follower pairs from the original NGSIM dataset (\citealp{thiemann2008estimating}; \citealp{kesting2008calibrating}; \citealp{montanino2013making}; \citealp{treiber2013traffic}). After the re-extraction of the NGSIM I-80 datasets, we use five car-following pairs with car-following period greater than 60s shown in Table \ref{tab: ngsim table} for evaluation. Same as the evaluation in Section \ref{sec:compare with other approaches}, the proposed two-stage DDPG follower is compared with the real follower in the dataset, pure DDPG follower, and IDM follower, where each IDM follower is calibrated according to the evaluation database.\par

The evaluation results illustrated in Figure \ref{fig:ngsim evaluation} and Table \ref{tab:ttc for ngsim} are similar to the results in Section \ref{sec:compare with other approaches}. From the velocity trajectories and bumper-to-bumper gap figures, we observe that the two-stage DDPG followers have a shorter reaction time than other followers, and follow closer to the leading vehicle for travel efficiency compared with the passive pure DDPG follower. The real follower and IDM follower show more aggressive risky driving behavior and tend to drive too close to the leader. The hazards of this kind of reckless driving are reflected in the TTC analysis. As shown in the TTC figures, the real followers and IDM followers have smaller TTC, in cases 3, 4, and 5 which have stop-and-go traffic situations, even with TTC smaller than 2s, which can be seen as dangerous situations. While the two-stage DDPG followers never have TTC smaller than 3.5s with smaller variances and drive safer than pure DDPG follower with higher driving efficiency. It is important to mention that the NGSIM I-80 database is not used during the training process, but only for evaluation.
\begin{table*}
	\centering
	\begin{threeparttable}
	\begin{tabular}{ccccc}
		\Xhline{1pt}
		Case & Leader ID  & Follower ID   & Start time & End time\\
		\hline
		1 & Vehicle 2159  & Vehicle 2159 & 654.0s & 715.0s\\
		2 & Vehicle 2914  & Vehicle 2923 & 803.0s & 868.0s\\
		3 & Vehicle 3006  & Vehicle 3020 & 799.0s & 861.0s\\
		4 & Vehicle 3021  & Vehicle 3031 & 823.0s & 883.0s\\
		5 & Vehicle 3051  & Vehicle 3053 & 830.0s & 899.0s\\
		\Xhline{1pt}
	\end{tabular}
	\caption{Five different car-following pairs are selected from the re-extracted NGSIM I-80 dataset.}
	\label{tab: ngsim table}
	\end{threeparttable}
\end{table*}

\begin{figure}
        \centering 
        \subfloat[Case 1: vehicle 2159 and vehicle 2166 as leader-follower pair (T = 654.0s$\sim$715.0s) ]{\includegraphics[width=\textwidth]{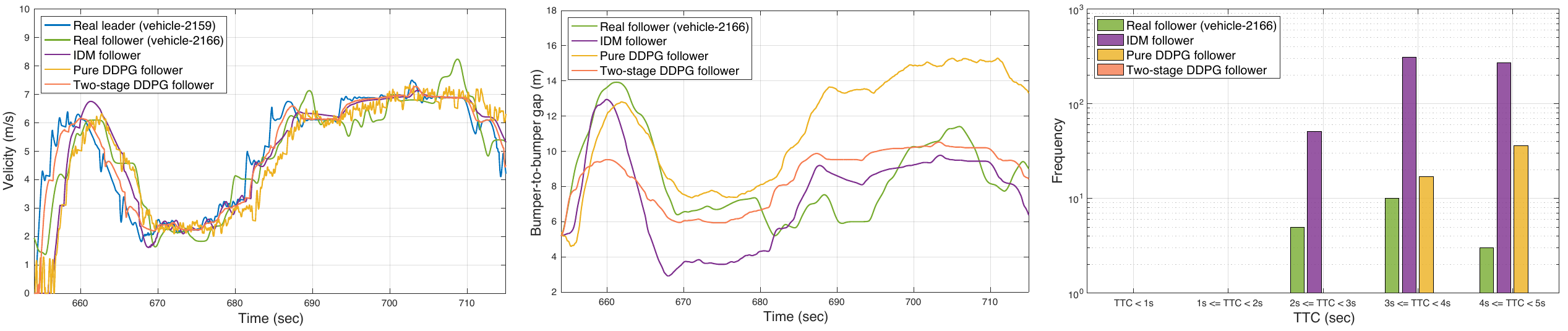}}
        \hfill
         \subfloat[Case 2: vehicle 2914 and vehicle 2923 as leader-follower pair (T = 803.0s$\sim$868.0s)]{
        \includegraphics[width=\textwidth]{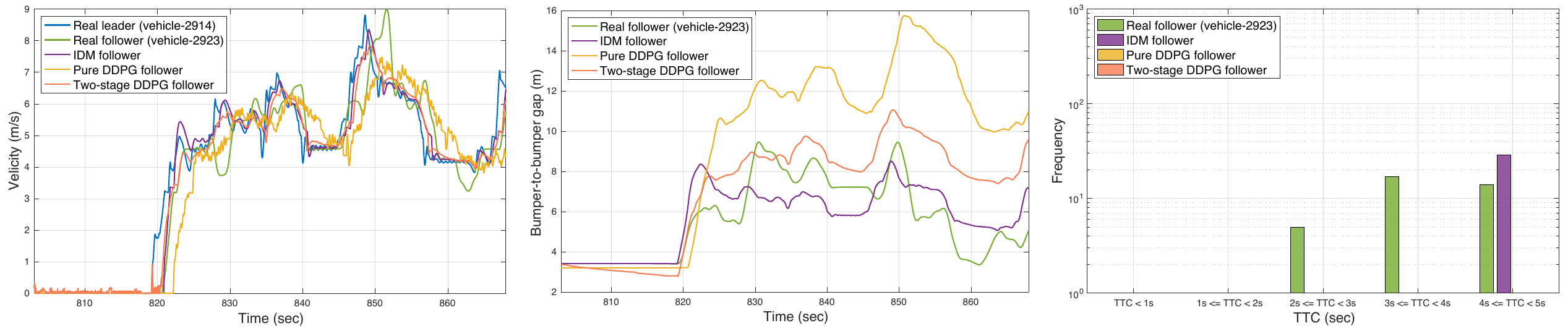}}
        \hfill
        \subfloat[Case 3: vehicle 3006 and vehicle 3020 as leader-follower pair (T = 799.0s$\sim$861.0s)]{
        \includegraphics[width=\textwidth]{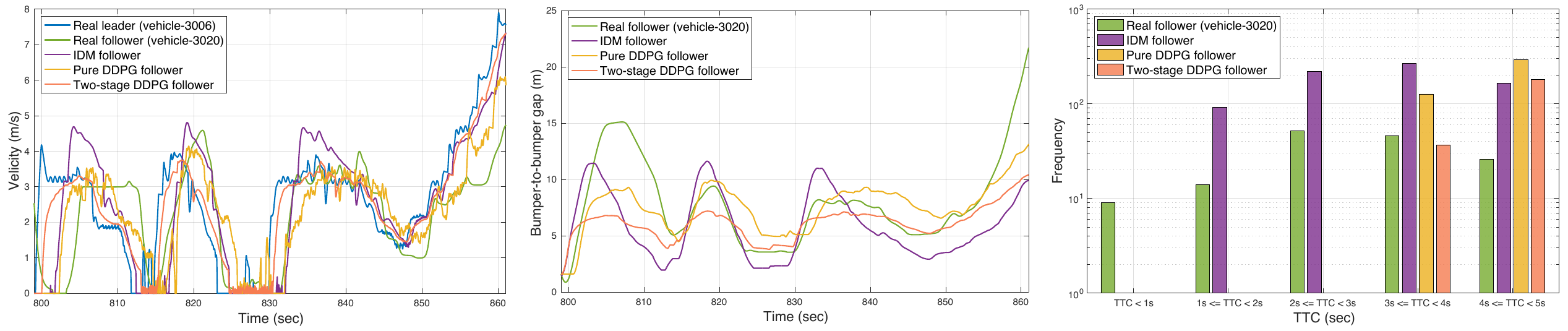}}
        \end{figure}
        \clearpage
        \begin{figure}
        \centering
        \ContinuedFloat
        \addtocounter{figure}{1}
        \subfloat[Case 4: vehicle 3021 and vehicle 3031 as leader-follower pair (T = 823.0s$\sim$883.0s)]{
        \includegraphics[width=\textwidth]{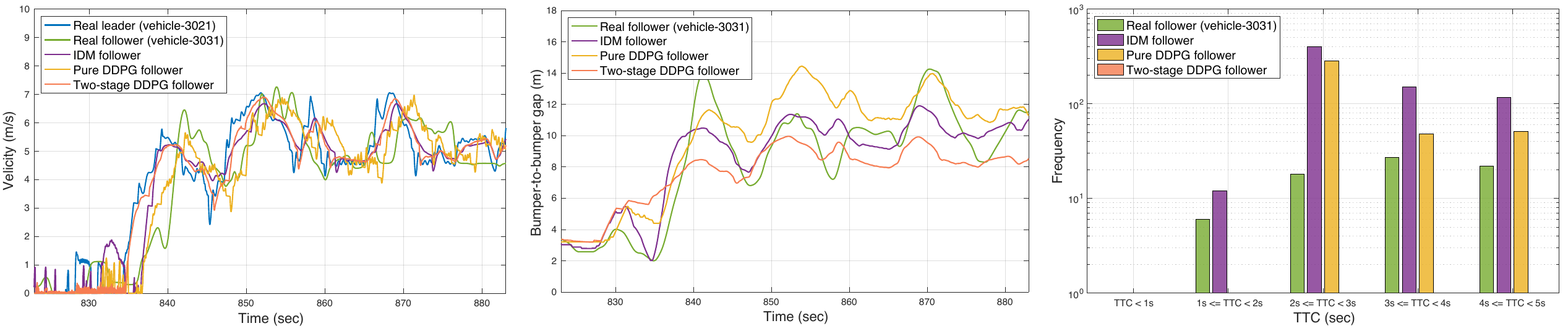}}
            \hfill
        \subfloat[Case 5: vehicle 3051 and vehicle 3053 as leader-follower pair (T = 830.0s$\sim$899.0s)]{
        \includegraphics[width=\textwidth]{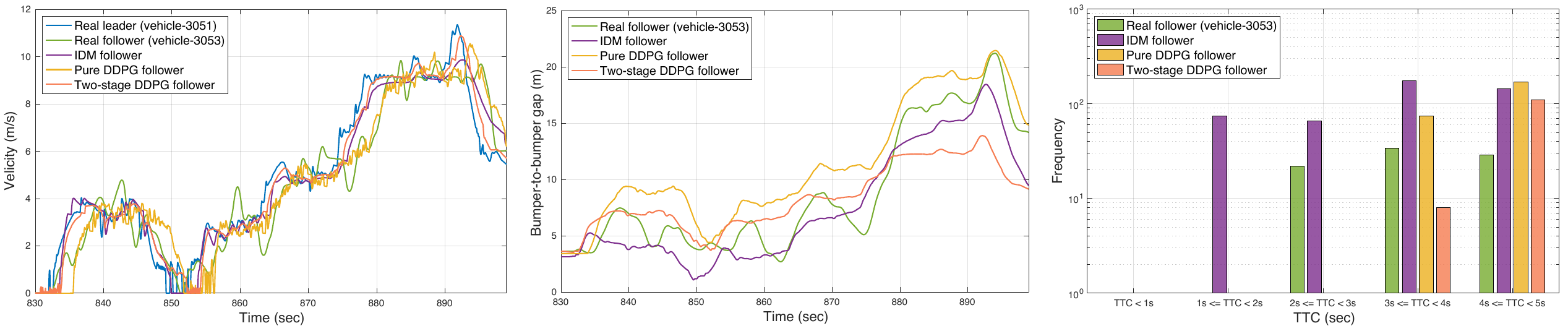}}
            \hfill
        \caption{We compare two-stage DDPG follower (red) with IDM follower (purple), pure DDPG follower (yellow), and real follower (green) following the real leader (blue) in the NGSIM I-80 dataset in CARLA. (a) $\sim$ (e) are the results following five different leader trajectories, from left to right are the velocity trajectories, bumper-to-bumper gap, and TTC distribution for different followers respectively.}
        \label{fig:ngsim evaluation}
\end{figure}

\begin{table*}
    \centering
    \begin{tabular}{llcccc}
    \Xhline{1pt}
      \multicolumn{2}{c}{\multirow{2}{*}{Agents}} & \multicolumn{4}{c}{TTC (Sec)}\\
    \cline{3-6}
     & & Minimum & Mean & Median & Std. dev\\
    \hline
    \multirow{4}{*}{Case 1}&Real driver & 2.86 & 6.92 & 7.56 & 2.16\\
    &IDM   & 2.36 & 5.64 & 5.18 & 2.00\\
    &Pure DDPG  & 3.61 & 7.55 & 7.88 & 1.74\\
	&Two-stage DDPG  & 5.89 & 8.41 & 8.62 & 1.29\\
	\hline
    \multirow{4}{*}{Case 2}&Real driver & 2.77 & 5.68 & 5.29 & 1.96\\
    &IDM   & 4.51 & 7.69 & 7.94 & 1.56\\
    &Pure DDPG  & 7.89 & 8.98 & 8.97 & 0.70\\
	&Two-stage DDPG  & 6.41 & 7.91 & 7.79 & 1.05\\
	\hline
    \multirow{4}{*}{Case 3}&Real driver & 0.75 & 5.28 & 5.50 & 2.41\\
    &IDM   & 1.80 & 6.43 & 7.12 & 2.40\\
    &Pure DDPG  & 3.11 & 6.13 & 5.79 & 1.89\\
	&Two-stage DDPG  & 3.51 & 6.81 & 6.25 & 1.94\\
	\hline
    \multirow{4}{*}{Case 4}&Real driver & 1.92 & 5.46 & 5.01 & 2.30\\
    &IDM   & 1.98 & 4.65 & 3.65 & 2.62\\
    &Pure DDPG  & 2.00 & 5.19 & 4.16 & 2.77\\
	&Two-stage DDPG  & 5.35 & 7.98 & 8.07 & 1.23\\
	\hline
    \multirow{4}{*}{Case 5}&Real driver & 2.33 & 5.71 & 5.24 & 2.32\\
    &IDM   & 1.19 & 5.91 & 6.41 & 2.56\\
    &Pure DDPG  & 3.06 & 6.48 & 6.09 & 2.25\\
	&Two-stage DDPG  & 3.90 & 7.31 & 7.36 & 2.25\\
	\Xhline{1pt}
    \end{tabular}
    \caption{Evaluation results of TTC under 10s for different agents in different evaluation cases from NGSIM dataset.}
    \label{tab:ttc for ngsim}
\end{table*}

\subsection{Evaluation with self-defined leading vehicle speed profile}
\begin{figure}
        \newcommand{\w}{0.45}
        \centering 
        \subfloat[Speed comparison of different follower]{\includegraphics[width=\textwidth]{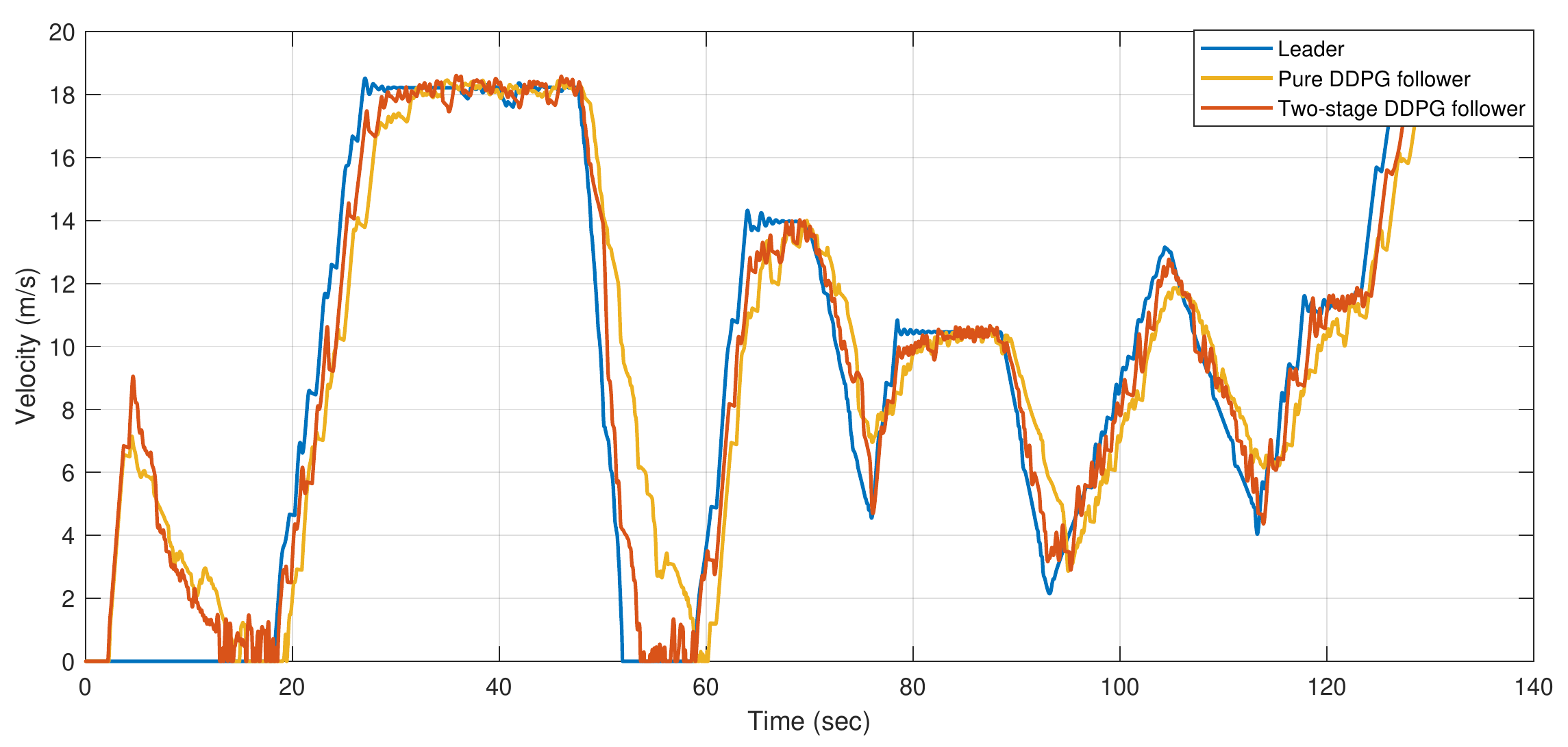}}
            \hfil 
         \subfloat[Bumper-to-bumper gap comparison of different follower]{
            \includegraphics[width=\textwidth]{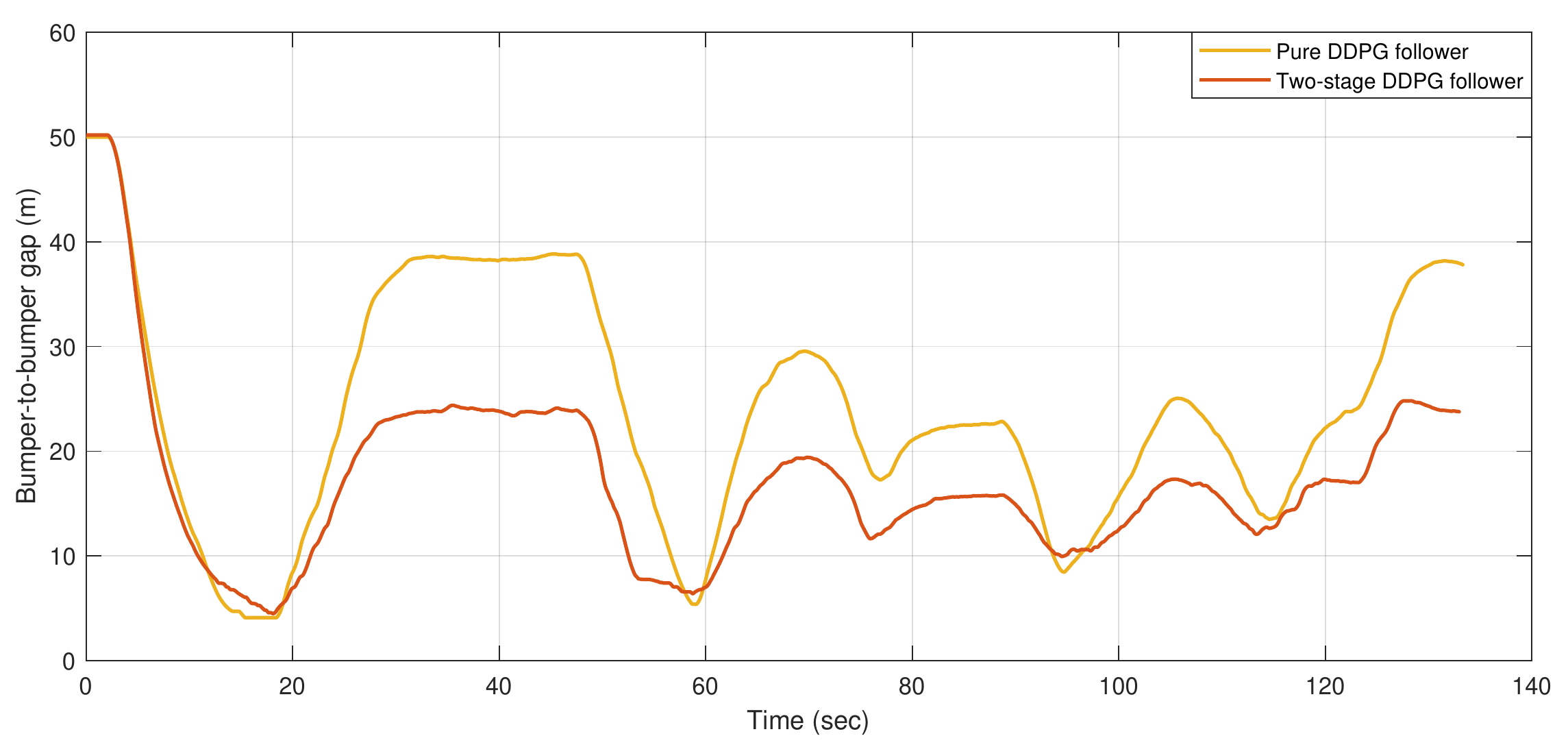}}
            \hfil
        \caption{The evaluation results by following an external leader trajectory for pure DDPG agent (yellow) and DDPG agent with human experience (red).} 
        \label{fig:stastic_leader}
\end{figure}

Next, we designed a leading vehicle speed profile that accounts for specific behavior not seen in the Napoli dataset and NGSIM dataset. First, we define a velocity trajectory for the leading vehicle with some safety-critical and common driving behavior. Figure \ref{fig:stastic_leader} illustrates the comparison of the performance of pure DDPG agent and two-stage DDPG agent.

The initial gap between followers and leader is 50 meters, and the initial velocity of each follower is \SI{0}{m/s}. Both followers accelerate from a standstill state as they realize the big distance to the leader. By approaching the leader, both followers reduce their velocity to obtain a safe distance. As the leading vehicle starts to accelerate from T = 18 s and reaches the velocity of \SI{18}{m/s}, both followers follow this change. The pure DDPG agent is clearly much passive, needs more reaction time, and keeps a bigger distance during the constant velocity driving stage. Due to learning from the human driving behavior, the two-stage DDPG follower has a shorter reaction time and keeps much closer to the leader, which improves the driving efficiency.\par
From T = 48 s, the leader performs safety-critical braking with the deceleration of \SI{-5}{m/s^2} until it stops. Our two-stage DDPG follower notices this braking timely and starts to decelerate accordingly while keeping a safe distance all the time. Same as the starting point, the pure DDPG follower can not sense this braking opportunely and leads to a smaller distance to the leader. Afterward, the leader drives with some common accelerate and decelerate scenarios, and the two-stage DDPG follower is able to keep a more appropriate distance than the pure DDPG follower.




\section{Conclusion and outlook}
\label{sec:Conclusion}
In this study, we proposed a so-called two-stage DDPG method to combine the real-world human driving dataset with DRL method to obtain a car-following agent which has human experience. With this approach, the agent can modify its driving policy by extracting the good behaviors from real-world human demonstrations. This will help the autonomous agent adapt to the real-world traffic environment. Moreover, we also proposed a framework that uses ROS and hyper-realistic autonomous driving simulator CARLA together to overcome the communication problem for car-following tasks. The agent can obtain all the information from other vehicles, for instance, position, velocity, acceleration, etc. With this framework. We used a small control network to connect to the policy network from DDPG for generalization. This control network can be retrained repeatedly for different types of vehicles whether in reality or simulation. Therefore, the trained DDPG networks can be reused many times. To the best of our knowledge, this is the first time that the CARLA-ROS communication method and the general control network are used for DRL in CARLA. \par
To evaluate the results, we designed different driving scenarios, analyzed the improvement of the two-stage DDPG agent qualitatively and quantitatively. The results showed that the two-stage DDPG agent was superior to the pure DDPG agent in terms of safety and driving efficiency. The proposed framework with CARLA-ROS communication made the car-following task with DRL in CARLA very easy to train and evaluate.\par
Although our research showed that combining the real-world human demonstrations improved the driving ability of the DDPG agent, there are still some suboptimal issues. One of them is that we calculate the reward for the actions in the real driving dataset based on the expert experience reward function. This cannot fully represent the behavior of the real driver, since this reward function is not the internal reward function when human drives in real-world traffic. A better solution is to use Inverse Reinforcement Learning (IRL) to obtain the internal reward function from the human driver. This will be discussed in the forthcoming paper.

\section*{Acknowledgement}
This work was funded by ScaDS.AI (Center for Scalable Data Analytics and Artificial Intelligence) Dresden/Leipzig. We would like to thank Martin Treiber, Martin Waltz, and Fabian Hart for constructive criticism of the manuscript, and also thank Vincenzo Punzo and Martin Treiber for providing the Napoli and re-extracted NGSIM I-80 dataset.



\bibliographystyle{elsarticle-harv} 
\bibliography{cas-refs}

\begin{thebibliography}{53}
\expandafter\ifx\csname natexlab\endcsname\relax\def\natexlab#1{#1}\fi
\providecommand{\url}[1]{\texttt{#1}}
\providecommand{\href}[2]{#2}
\providecommand{\path}[1]{#1}
\providecommand{\DOIprefix}{doi:}
\providecommand{\ArXivprefix}{arXiv:}
\providecommand{\URLprefix}{URL: }
\providecommand{\Pubmedprefix}{pmid:}
\providecommand{\doi}[1]{\href{http://dx.doi.org/#1}{\path{#1}}}
\providecommand{\Pubmed}[1]{\href{pmid:#1}{\path{#1}}}
\providecommand{\bibinfo}[2]{#2}
\ifx\xfnm\relax \def\xfnm[#1]{\unskip,\space#1}\fi
\bibitem[{Akhloufi et~al.(2019)Akhloufi, Arola and Bonnet}]{akhloufi2019drones}
\bibinfo{author}{Akhloufi, M.A.}, \bibinfo{author}{Arola, S.},
  \bibinfo{author}{Bonnet, A.}, \bibinfo{year}{2019}.
\newblock \bibinfo{title}{Drones chasing drones: Reinforcement learning and
  deep search area proposal}.
\newblock \bibinfo{journal}{Drones} \bibinfo{volume}{3}, \bibinfo{pages}{58}.
\bibitem[{Bojarski et~al.(2016)Bojarski, Del~Testa, Dworakowski, Firner, Flepp,
  Goyal, Jackel, Monfort, Muller, Zhang et~al.}]{bojarski2016end}
\bibinfo{author}{Bojarski, M.}, \bibinfo{author}{Del~Testa, D.},
  \bibinfo{author}{Dworakowski, D.}, \bibinfo{author}{Firner, B.},
  \bibinfo{author}{Flepp, B.}, \bibinfo{author}{Goyal, P.},
  \bibinfo{author}{Jackel, L.D.}, \bibinfo{author}{Monfort, M.},
  \bibinfo{author}{Muller, U.}, \bibinfo{author}{Zhang, J.}, et~al.,
  \bibinfo{year}{2016}.
\newblock \bibinfo{title}{End to end learning for self-driving cars}.
\newblock \bibinfo{journal}{arXiv preprint arXiv:1604.07316} .
\bibitem[{Codevilla et~al.(2018)Codevilla, M{\"u}ller, L{\'o}pez, Koltun and
  Dosovitskiy}]{codevilla2018end}
\bibinfo{author}{Codevilla, F.}, \bibinfo{author}{M{\"u}ller, M.},
  \bibinfo{author}{L{\'o}pez, A.}, \bibinfo{author}{Koltun, V.},
  \bibinfo{author}{Dosovitskiy, A.}, \bibinfo{year}{2018}.
\newblock \bibinfo{title}{End-to-end driving via conditional imitation
  learning}, in: \bibinfo{booktitle}{2018 IEEE International Conference on
  Robotics and Automation (ICRA)}, \bibinfo{organization}{IEEE}. pp.
  \bibinfo{pages}{4693--4700}.
\bibitem[{Codevilla et~al.(2019)Codevilla, Santana, L{\'o}pez and
  Gaidon}]{codevilla2019exploring}
\bibinfo{author}{Codevilla, F.}, \bibinfo{author}{Santana, E.},
  \bibinfo{author}{L{\'o}pez, A.M.}, \bibinfo{author}{Gaidon, A.},
  \bibinfo{year}{2019}.
\newblock \bibinfo{title}{Exploring the limitations of behavior cloning for
  autonomous driving}, in: \bibinfo{booktitle}{Proceedings of the IEEE/CVF
  International Conference on Computer Vision}, pp.
  \bibinfo{pages}{9329--9338}.
\bibitem[{Dadashi et~al.(2021)Dadashi, Rezaeifar, Vieillard, Hussenot, Pietquin
  and Geist}]{dadashi2021offline}
\bibinfo{author}{Dadashi, R.}, \bibinfo{author}{Rezaeifar, S.},
  \bibinfo{author}{Vieillard, N.}, \bibinfo{author}{Hussenot, L.},
  \bibinfo{author}{Pietquin, O.}, \bibinfo{author}{Geist, M.},
  \bibinfo{year}{2021}.
\newblock \bibinfo{title}{Offline reinforcement learning with pseudometric
  learning}, in: \bibinfo{booktitle}{International Conference on Machine
  Learning}, \bibinfo{organization}{PMLR}. pp. \bibinfo{pages}{2307--2318}.
\bibitem[{Dosovitskiy et~al.(2017)Dosovitskiy, Ros, Codevilla, Lopez and
  Koltun}]{Dosovitskiy17}
\bibinfo{author}{Dosovitskiy, A.}, \bibinfo{author}{Ros, G.},
  \bibinfo{author}{Codevilla, F.}, \bibinfo{author}{Lopez, A.},
  \bibinfo{author}{Koltun, V.}, \bibinfo{year}{2017}.
\newblock \bibinfo{title}{{CARLA}: {An} open urban driving simulator}, in:
  \bibinfo{booktitle}{Proceedings of the 1st Annual Conference on Robot
  Learning}, pp. \bibinfo{pages}{1--16}.
\bibitem[{Dupuis et~al.(2010)Dupuis, Strobl and
  Grezlikowski}]{dupuis2010opendrive}
\bibinfo{author}{Dupuis, M.}, \bibinfo{author}{Strobl, M.},
  \bibinfo{author}{Grezlikowski, H.}, \bibinfo{year}{2010}.
\newblock \bibinfo{title}{Opendrive 2010 and beyond--status and future of the
  de facto standard for the description of road networks}, in:
  \bibinfo{booktitle}{Proc. of the Driving Simulation Conference Europe}, pp.
  \bibinfo{pages}{231--242}.
\bibitem[{Dworak et~al.(2019)Dworak, Ciepiela, Derbisz, Izzat, Komorkiewicz and
  W{\'o}jcik}]{dworak2019performance}
\bibinfo{author}{Dworak, D.}, \bibinfo{author}{Ciepiela, F.},
  \bibinfo{author}{Derbisz, J.}, \bibinfo{author}{Izzat, I.},
  \bibinfo{author}{Komorkiewicz, M.}, \bibinfo{author}{W{\'o}jcik, M.},
  \bibinfo{year}{2019}.
\newblock \bibinfo{title}{Performance of lidar object detection deep learning
  architectures based on artificially generated point cloud data from carla
  simulator}, in: \bibinfo{booktitle}{2019 24th International Conference on
  Methods and Models in Automation and Robotics (MMAR)},
  \bibinfo{organization}{IEEE}. pp. \bibinfo{pages}{600--605}.
\bibitem[{Fujimoto et~al.(2019)Fujimoto, Meger and Precup}]{fujimoto2019off}
\bibinfo{author}{Fujimoto, S.}, \bibinfo{author}{Meger, D.},
  \bibinfo{author}{Precup, D.}, \bibinfo{year}{2019}.
\newblock \bibinfo{title}{Off-policy deep reinforcement learning without
  exploration}, in: \bibinfo{booktitle}{International Conference on Machine
  Learning}, \bibinfo{organization}{PMLR}. pp. \bibinfo{pages}{2052--2062}.
\bibitem[{Gazis et~al.(1961)Gazis, Herman and Rothery}]{gazis1961nonlinear}
\bibinfo{author}{Gazis, D.C.}, \bibinfo{author}{Herman, R.},
  \bibinfo{author}{Rothery, R.W.}, \bibinfo{year}{1961}.
\newblock \bibinfo{title}{Nonlinear follow-the-leader models of traffic flow}.
\newblock \bibinfo{journal}{Operations research} \bibinfo{volume}{9},
  \bibinfo{pages}{545--567}.
\bibitem[{G{\'o}mez-Hu{\'e}lamo et~al.(2020)G{\'o}mez-Hu{\'e}lamo, Del~Egido,
  Bergasa, Barea, L{\'o}pez-Guill{\'e}n, Arango, Araluce and
  L{\'o}pez}]{gomez2020train}
\bibinfo{author}{G{\'o}mez-Hu{\'e}lamo, C.}, \bibinfo{author}{Del~Egido, J.},
  \bibinfo{author}{Bergasa, L.M.}, \bibinfo{author}{Barea, R.},
  \bibinfo{author}{L{\'o}pez-Guill{\'e}n, E.}, \bibinfo{author}{Arango, F.},
  \bibinfo{author}{Araluce, J.}, \bibinfo{author}{L{\'o}pez, J.},
  \bibinfo{year}{2020}.
\newblock \bibinfo{title}{Train here, drive there: Simulating real-world use
  cases with fully-autonomous driving architecture in carla simulator}, in:
  \bibinfo{booktitle}{Workshop of Physical Agents},
  \bibinfo{organization}{Springer}. pp. \bibinfo{pages}{44--59}.
\bibitem[{Gu et~al.(2017)Gu, Holly, Lillicrap and Levine}]{gu2017deep}
\bibinfo{author}{Gu, S.}, \bibinfo{author}{Holly, E.},
  \bibinfo{author}{Lillicrap, T.}, \bibinfo{author}{Levine, S.},
  \bibinfo{year}{2017}.
\newblock \bibinfo{title}{Deep reinforcement learning for robotic manipulation
  with asynchronous off-policy updates}, in: \bibinfo{booktitle}{2017 IEEE
  international conference on robotics and automation (ICRA)},
  \bibinfo{organization}{IEEE}. pp. \bibinfo{pages}{3389--3396}.
\bibitem[{de~Haan et~al.(2019)de~Haan, Jayaraman and Levine}]{de2019causal}
\bibinfo{author}{de~Haan, P.}, \bibinfo{author}{Jayaraman, D.},
  \bibinfo{author}{Levine, S.}, \bibinfo{year}{2019}.
\newblock \bibinfo{title}{Causal confusion in imitation learning}.
\newblock \bibinfo{journal}{Advances in Neural Information Processing Systems}
  \bibinfo{volume}{32}, \bibinfo{pages}{11698--11709}.
\bibitem[{Hanin and Rolnick(2018)}]{hanin2018start}
\bibinfo{author}{Hanin, B.}, \bibinfo{author}{Rolnick, D.},
  \bibinfo{year}{2018}.
\newblock \bibinfo{title}{How to start training: The effect of initialization
  and architecture}.
\newblock \bibinfo{journal}{arXiv preprint arXiv:1803.01719} .
\bibitem[{Hart et~al.(2021)Hart, Okhrin and Treiber}]{hart2021formulation}
\bibinfo{author}{Hart, F.}, \bibinfo{author}{Okhrin, O.},
  \bibinfo{author}{Treiber, M.}, \bibinfo{year}{2021}.
\newblock \bibinfo{title}{Formulation and validation of a car-following model
  based on deep reinforcement learning}.
\newblock \bibinfo{journal}{arXiv preprint arXiv:2109.14268} .
\bibitem[{Hayward(1972)}]{Hayward1972NEARMISSDT}
\bibinfo{author}{Hayward, J.C.}, \bibinfo{year}{1972}.
\newblock \bibinfo{title}{Near-miss determination through use of a scale of
  danger}.
\newblock \bibinfo{journal}{Highway Research Record} .
\bibitem[{He et~al.(2017)He, Zhao and Yin}]{he2017integrated}
\bibinfo{author}{He, Y.}, \bibinfo{author}{Zhao, N.}, \bibinfo{author}{Yin,
  H.}, \bibinfo{year}{2017}.
\newblock \bibinfo{title}{Integrated networking, caching, and computing for
  connected vehicles: A deep reinforcement learning approach}.
\newblock \bibinfo{journal}{IEEE Transactions on Vehicular Technology}
  \bibinfo{volume}{67}, \bibinfo{pages}{44--55}.
\bibitem[{Hellmund et~al.(2016)Hellmund, Wirges, Ta{\c{s}}, Bandera and
  Salscheider}]{hellmund2016robot}
\bibinfo{author}{Hellmund, A.M.}, \bibinfo{author}{Wirges, S.},
  \bibinfo{author}{Ta{\c{s}}, {\"O}.{\c{S}}.}, \bibinfo{author}{Bandera, C.},
  \bibinfo{author}{Salscheider, N.O.}, \bibinfo{year}{2016}.
\newblock \bibinfo{title}{Robot operating system: A modular software framework
  for automated driving}, in: \bibinfo{booktitle}{2016 IEEE 19th International
  Conference on Intelligent Transportation Systems (ITSC)},
  \bibinfo{organization}{IEEE}. pp. \bibinfo{pages}{1564--1570}.
\bibitem[{Hester et~al.(2018)Hester, Vecerik, Pietquin, Lanctot, Schaul, Piot,
  Horgan, Quan, Sendonaris, Osband et~al.}]{hester2018deep}
\bibinfo{author}{Hester, T.}, \bibinfo{author}{Vecerik, M.},
  \bibinfo{author}{Pietquin, O.}, \bibinfo{author}{Lanctot, M.},
  \bibinfo{author}{Schaul, T.}, \bibinfo{author}{Piot, B.},
  \bibinfo{author}{Horgan, D.}, \bibinfo{author}{Quan, J.},
  \bibinfo{author}{Sendonaris, A.}, \bibinfo{author}{Osband, I.}, et~al.,
  \bibinfo{year}{2018}.
\newblock \bibinfo{title}{Deep q-learning from demonstrations}, in:
  \bibinfo{booktitle}{Proceedings of the AAAI Conference on Artificial
  Intelligence}, p. \bibinfo{pages}{3223–3230}.
\bibitem[{Huang et~al.(2021)Huang, Wu and Lv}]{huang2021efficient}
\bibinfo{author}{Huang, Z.}, \bibinfo{author}{Wu, J.}, \bibinfo{author}{Lv,
  C.}, \bibinfo{year}{2021}.
\newblock \bibinfo{title}{Efficient deep reinforcement learning with imitative
  expert priors for autonomous driving}.
\newblock \bibinfo{journal}{arXiv preprint arXiv:2103.10690} .
\bibitem[{Isele et~al.(2018)Isele, Rahimi, Cosgun, Subramanian and
  Fujimura}]{isele2018navigating}
\bibinfo{author}{Isele, D.}, \bibinfo{author}{Rahimi, R.},
  \bibinfo{author}{Cosgun, A.}, \bibinfo{author}{Subramanian, K.},
  \bibinfo{author}{Fujimura, K.}, \bibinfo{year}{2018}.
\newblock \bibinfo{title}{Navigating occluded intersections with autonomous
  vehicles using deep reinforcement learning}, in: \bibinfo{booktitle}{2018
  IEEE International Conference on Robotics and Automation (ICRA)},
  \bibinfo{organization}{IEEE}. pp. \bibinfo{pages}{2034--2039}.
\bibitem[{Kalman(1960)}]{kalman1960new}
\bibinfo{author}{Kalman, R.E.}, \bibinfo{year}{1960}.
\newblock \bibinfo{title}{A new approach to linear filtering and prediction
  problems}.
\newblock \bibinfo{journal}{Transaction of the ASME — Journal of Basic
  Engineering} , \bibinfo{pages}{35--45}.
\bibitem[{Kesting and Treiber(2008)}]{kesting2008calibrating}
\bibinfo{author}{Kesting, A.}, \bibinfo{author}{Treiber, M.},
  \bibinfo{year}{2008}.
\newblock \bibinfo{title}{Calibrating car-following models by using trajectory
  data: Methodological study}.
\newblock \bibinfo{journal}{Transportation Research Record}
  \bibinfo{volume}{2088}, \bibinfo{pages}{148--156}.
\bibitem[{Kikuchi and Chakroborty(1992)}]{kikuchi1992car}
\bibinfo{author}{Kikuchi, S.}, \bibinfo{author}{Chakroborty, P.},
  \bibinfo{year}{1992}.
\newblock \bibinfo{title}{Car-following model based on fuzzy inference system}.
\newblock \bibinfo{journal}{Transportation Research Record} ,
  \bibinfo{pages}{82--82}.
\bibitem[{Kim(1999)}]{kim1999normalization}
\bibinfo{author}{Kim, D.}, \bibinfo{year}{1999}.
\newblock \bibinfo{title}{Normalization methods for input and output vectors in
  backpropagation neural networks}.
\newblock \bibinfo{journal}{International journal of computer mathematics}
  \bibinfo{volume}{71}, \bibinfo{pages}{161--171}.
\bibitem[{Levine et~al.(2020)Levine, Kumar, Tucker and Fu}]{levine2020offline}
\bibinfo{author}{Levine, S.}, \bibinfo{author}{Kumar, A.},
  \bibinfo{author}{Tucker, G.}, \bibinfo{author}{Fu, J.}, \bibinfo{year}{2020}.
\newblock \bibinfo{title}{Offline reinforcement learning: Tutorial, review, and
  perspectives on open problems}.
\newblock \bibinfo{journal}{arXiv preprint arXiv:2005.01643} .
\bibitem[{Lillicrap et~al.(2015)Lillicrap, Hunt, Pritzel, Heess, Erez, Tassa,
  Silver and Wierstra}]{lillicrap2015continuous}
\bibinfo{author}{Lillicrap, T.P.}, \bibinfo{author}{Hunt, J.J.},
  \bibinfo{author}{Pritzel, A.}, \bibinfo{author}{Heess, N.},
  \bibinfo{author}{Erez, T.}, \bibinfo{author}{Tassa, Y.},
  \bibinfo{author}{Silver, D.}, \bibinfo{author}{Wierstra, D.},
  \bibinfo{year}{2015}.
\newblock \bibinfo{title}{Continuous control with deep reinforcement learning}.
\newblock \bibinfo{journal}{arXiv preprint arXiv:1509.02971} .
\bibitem[{Lin(1992)}]{lin1992self}
\bibinfo{author}{Lin, L.J.}, \bibinfo{year}{1992}.
\newblock \bibinfo{title}{Self-improving reactive agents based on reinforcement
  learning, planning and teaching}.
\newblock \bibinfo{journal}{Machine learning} \bibinfo{volume}{8},
  \bibinfo{pages}{293--321}.
\bibitem[{Litman(2017)}]{litman2017autonomous}
\bibinfo{author}{Litman, T.}, \bibinfo{year}{2017}.
\newblock \bibinfo{title}{Autonomous vehicle implementation predictions}.
\newblock \bibinfo{publisher}{Victoria Transport Policy Institute Victoria,
  Canada}.
\bibitem[{Liu et~al.(2021)Liu, Huang and Lv}]{liu2021improved}
\bibinfo{author}{Liu, H.}, \bibinfo{author}{Huang, Z.}, \bibinfo{author}{Lv,
  C.}, \bibinfo{year}{2021}.
\newblock \bibinfo{title}{Improved deep reinforcement learning with expert
  demonstrations for urban autonomous driving}.
\newblock \bibinfo{journal}{arXiv preprint arXiv:2102.09243} .
\bibitem[{Minderhoud and Bovy(2001)}]{minderhoud2001extended}
\bibinfo{author}{Minderhoud, M.M.}, \bibinfo{author}{Bovy, P.H.},
  \bibinfo{year}{2001}.
\newblock \bibinfo{title}{Extended time-to-collision measures for road traffic
  safety assessment}.
\newblock \bibinfo{journal}{Accident Analysis \& Prevention}
  \bibinfo{volume}{33}, \bibinfo{pages}{89--97}.
\bibitem[{Mnih et~al.(2013)Mnih, Kavukcuoglu, Silver, Graves, Antonoglou,
  Wierstra and Riedmiller}]{mnih2013playing}
\bibinfo{author}{Mnih, V.}, \bibinfo{author}{Kavukcuoglu, K.},
  \bibinfo{author}{Silver, D.}, \bibinfo{author}{Graves, A.},
  \bibinfo{author}{Antonoglou, I.}, \bibinfo{author}{Wierstra, D.},
  \bibinfo{author}{Riedmiller, M.}, \bibinfo{year}{2013}.
\newblock \bibinfo{title}{Playing atari with deep reinforcement learning}.
\newblock \bibinfo{journal}{arXiv preprint arXiv:1312.5602} .
\bibitem[{Mnih et~al.(2015)Mnih, Kavukcuoglu, Silver, Rusu, Veness, Bellemare,
  Graves, Riedmiller, Fidjeland, Ostrovski et~al.}]{mnih2015human}
\bibinfo{author}{Mnih, V.}, \bibinfo{author}{Kavukcuoglu, K.},
  \bibinfo{author}{Silver, D.}, \bibinfo{author}{Rusu, A.A.},
  \bibinfo{author}{Veness, J.}, \bibinfo{author}{Bellemare, M.G.},
  \bibinfo{author}{Graves, A.}, \bibinfo{author}{Riedmiller, M.},
  \bibinfo{author}{Fidjeland, A.K.}, \bibinfo{author}{Ostrovski, G.}, et~al.,
  \bibinfo{year}{2015}.
\newblock \bibinfo{title}{Human-level control through deep reinforcement
  learning}.
\newblock \bibinfo{journal}{nature} \bibinfo{volume}{518},
  \bibinfo{pages}{529--533}.
\bibitem[{Montanino and Punzo(2013)}]{montanino2013making}
\bibinfo{author}{Montanino, M.}, \bibinfo{author}{Punzo, V.},
  \bibinfo{year}{2013}.
\newblock \bibinfo{title}{Making ngsim data usable for studies on traffic flow
  theory: Multistep method for vehicle trajectory reconstruction}.
\newblock \bibinfo{journal}{Transportation Research Record}
  \bibinfo{volume}{2390}, \bibinfo{pages}{99--111}.
\bibitem[{Ngai and Yung(2011)}]{ngai2011multiple}
\bibinfo{author}{Ngai, D.C.K.}, \bibinfo{author}{Yung, N.H.C.},
  \bibinfo{year}{2011}.
\newblock \bibinfo{title}{A multiple-goal reinforcement learning method for
  complex vehicle overtaking maneuvers}.
\newblock \bibinfo{journal}{IEEE Transactions on Intelligent Transportation
  Systems} \bibinfo{volume}{12}, \bibinfo{pages}{509--522}.
\bibitem[{Niranjan et~al.(2021)Niranjan, VinayKarthik
  et~al.}]{niranjan2021deep}
\bibinfo{author}{Niranjan, D.}, \bibinfo{author}{VinayKarthik, B.}, et~al.,
  \bibinfo{year}{2021}.
\newblock \bibinfo{title}{Deep learning based object detection model for
  autonomous driving research using carla simulator}, in:
  \bibinfo{booktitle}{2021 2nd International Conference on Smart Electronics
  and Communication (ICOSEC)}, \bibinfo{organization}{IEEE}. pp.
  \bibinfo{pages}{1251--1258}.
\bibitem[{Nosrati et~al.(2018)Nosrati, Abolfathi, Elmahgiubi, Yadmellat, Luo,
  Zhang, Yao, Zhang and Jamil}]{nosrati2018towards}
\bibinfo{author}{Nosrati, M.S.}, \bibinfo{author}{Abolfathi, E.A.},
  \bibinfo{author}{Elmahgiubi, M.}, \bibinfo{author}{Yadmellat, P.},
  \bibinfo{author}{Luo, J.}, \bibinfo{author}{Zhang, Y.}, \bibinfo{author}{Yao,
  H.}, \bibinfo{author}{Zhang, H.}, \bibinfo{author}{Jamil, A.},
  \bibinfo{year}{2018}.
\newblock \bibinfo{title}{Towards practical hierarchical reinforcement learning
  for multi-lane autonomous driving}, in: \bibinfo{booktitle}{2018 NIPS MLITS
  Workshop}.
\bibitem[{Ornstein and Uhlenbeck(1930)}]{uhlenbeck1930theory}
\bibinfo{author}{Ornstein, L.S.}, \bibinfo{author}{Uhlenbeck, G.E.},
  \bibinfo{year}{1930}.
\newblock \bibinfo{title}{On the theory of the brownian motion}.
\newblock \bibinfo{journal}{Physical review} \bibinfo{volume}{36},
  \bibinfo{pages}{823}.
\bibitem[{Punzo et~al.(2005)Punzo, Formisano and
  Torrieri}]{punzo2005nonstationary}
\bibinfo{author}{Punzo, V.}, \bibinfo{author}{Formisano, D.J.},
  \bibinfo{author}{Torrieri, V.}, \bibinfo{year}{2005}.
\newblock \bibinfo{title}{Nonstationary kalman filter for estimation of
  accurate and consistent car-following data}.
\newblock \bibinfo{journal}{Transportation research record}
  \bibinfo{volume}{1934}, \bibinfo{pages}{2--12}.
\bibitem[{Quigley et~al.(2009)Quigley, Conley, Gerkey, Faust, Foote, Leibs,
  Wheeler, Ng et~al.}]{quigley2009ros}
\bibinfo{author}{Quigley, M.}, \bibinfo{author}{Conley, K.},
  \bibinfo{author}{Gerkey, B.}, \bibinfo{author}{Faust, J.},
  \bibinfo{author}{Foote, T.}, \bibinfo{author}{Leibs, J.},
  \bibinfo{author}{Wheeler, R.}, \bibinfo{author}{Ng, A.Y.}, et~al.,
  \bibinfo{year}{2009}.
\newblock \bibinfo{title}{Ros: an open-source robot operating system}, in:
  \bibinfo{booktitle}{ICRA workshop on open source software},
  \bibinfo{organization}{Kobe, Japan}. p.~\bibinfo{pages}{5}.
\bibitem[{Ravi~Kiran et~al.(2018)Ravi~Kiran, Roldao, Irastorza, Verastegui,
  Suss, Yogamani, Talpaert, Lepoutre and Trehard}]{ravi2018real}
\bibinfo{author}{Ravi~Kiran, B.}, \bibinfo{author}{Roldao, L.},
  \bibinfo{author}{Irastorza, B.}, \bibinfo{author}{Verastegui, R.},
  \bibinfo{author}{Suss, S.}, \bibinfo{author}{Yogamani, S.},
  \bibinfo{author}{Talpaert, V.}, \bibinfo{author}{Lepoutre, A.},
  \bibinfo{author}{Trehard, G.}, \bibinfo{year}{2018}.
\newblock \bibinfo{title}{Real-time dynamic object detection for autonomous
  driving using prior 3d-maps}, in: \bibinfo{booktitle}{Proceedings of the
  European Conference on Computer Vision (ECCV) Workshops}, pp.
  \bibinfo{pages}{0--0}.
\bibitem[{Sallab et~al.(2016)Sallab, Abdou, Perot and Yogamani}]{sallab2016end}
\bibinfo{author}{Sallab, A.E.}, \bibinfo{author}{Abdou, M.},
  \bibinfo{author}{Perot, E.}, \bibinfo{author}{Yogamani, S.},
  \bibinfo{year}{2016}.
\newblock \bibinfo{title}{End-to-end deep reinforcement learning for lane
  keeping assist}, in: \bibinfo{booktitle}{Proceedings of MLITS, NIPS
  Workshop}, pp. \bibinfo{pages}{vol. 2, pp. 1–9.}
\bibitem[{Silver et~al.(2016)Silver, Huang, Maddison, Guez, Sifre, Van
  Den~Driessche, Schrittwieser, Antonoglou, Panneershelvam, Lanctot
  et~al.}]{silver2016mastering}
\bibinfo{author}{Silver, D.}, \bibinfo{author}{Huang, A.},
  \bibinfo{author}{Maddison, C.J.}, \bibinfo{author}{Guez, A.},
  \bibinfo{author}{Sifre, L.}, \bibinfo{author}{Van Den~Driessche, G.},
  \bibinfo{author}{Schrittwieser, J.}, \bibinfo{author}{Antonoglou, I.},
  \bibinfo{author}{Panneershelvam, V.}, \bibinfo{author}{Lanctot, M.}, et~al.,
  \bibinfo{year}{2016}.
\newblock \bibinfo{title}{Mastering the game of go with deep neural networks
  and tree search}.
\newblock \bibinfo{journal}{nature} \bibinfo{volume}{529},
  \bibinfo{pages}{484--489}.
\bibitem[{Silver et~al.(2017)Silver, Hubert, Schrittwieser, Antonoglou, Lai,
  Guez, Lanctot, Sifre, Kumaran, Graepel et~al.}]{silver2017mastering}
\bibinfo{author}{Silver, D.}, \bibinfo{author}{Hubert, T.},
  \bibinfo{author}{Schrittwieser, J.}, \bibinfo{author}{Antonoglou, I.},
  \bibinfo{author}{Lai, M.}, \bibinfo{author}{Guez, A.},
  \bibinfo{author}{Lanctot, M.}, \bibinfo{author}{Sifre, L.},
  \bibinfo{author}{Kumaran, D.}, \bibinfo{author}{Graepel, T.}, et~al.,
  \bibinfo{year}{2017}.
\newblock \bibinfo{title}{Mastering chess and shogi by self-play with a general
  reinforcement learning algorithm}.
\newblock \bibinfo{journal}{arXiv preprint arXiv:1712.01815} .
\bibitem[{Thiemann et~al.(2008)Thiemann, Treiber and
  Kesting}]{thiemann2008estimating}
\bibinfo{author}{Thiemann, C.}, \bibinfo{author}{Treiber, M.},
  \bibinfo{author}{Kesting, A.}, \bibinfo{year}{2008}.
\newblock \bibinfo{title}{Estimating acceleration and lane-changing dynamics
  from next generation simulation trajectory data}.
\newblock \bibinfo{journal}{Transportation Research Record}
  \bibinfo{volume}{2088}, \bibinfo{pages}{90--101}.
\bibitem[{Thrun et~al.(2006)Thrun, Montemerlo, Dahlkamp, Stavens, Aron, Diebel,
  Fong, Gale, Halpenny, Hoffmann et~al.}]{thrun2006stanley}
\bibinfo{author}{Thrun, S.}, \bibinfo{author}{Montemerlo, M.},
  \bibinfo{author}{Dahlkamp, H.}, \bibinfo{author}{Stavens, D.},
  \bibinfo{author}{Aron, A.}, \bibinfo{author}{Diebel, J.},
  \bibinfo{author}{Fong, P.}, \bibinfo{author}{Gale, J.},
  \bibinfo{author}{Halpenny, M.}, \bibinfo{author}{Hoffmann, G.}, et~al.,
  \bibinfo{year}{2006}.
\newblock \bibinfo{title}{Stanley: The robot that won the darpa grand
  challenge}.
\newblock \bibinfo{journal}{Journal of field Robotics} \bibinfo{volume}{23},
  \bibinfo{pages}{661--692}.
\bibitem[{Tran and Le(2019)}]{tran2019robust}
\bibinfo{author}{Tran, L.A.}, \bibinfo{author}{Le, M.H.}, \bibinfo{year}{2019}.
\newblock \bibinfo{title}{Robust u-net-based road lane markings detection for
  autonomous driving}, in: \bibinfo{booktitle}{2019 International Conference on
  System Science and Engineering (ICSSE)}, \bibinfo{organization}{IEEE}. pp.
  \bibinfo{pages}{62--66}.
\bibitem[{Treiber et~al.(2000)Treiber, Hennecke and
  Helbing}]{treiber2000congested}
\bibinfo{author}{Treiber, M.}, \bibinfo{author}{Hennecke, A.},
  \bibinfo{author}{Helbing, D.}, \bibinfo{year}{2000}.
\newblock \bibinfo{title}{Congested traffic states in empirical observations
  and microscopic simulations}.
\newblock \bibinfo{journal}{Physical review E} \bibinfo{volume}{62},
  \bibinfo{pages}{1805}.
\bibitem[{Treiber and Kesting(2013)}]{treiber2013traffic}
\bibinfo{author}{Treiber, M.}, \bibinfo{author}{Kesting, A.},
  \bibinfo{year}{2013}.
\newblock \bibinfo{title}{Traffic flow dynamics}.
\newblock \bibinfo{journal}{Traffic Flow Dynamics: Data, Models and Simulation,
  Springer-Verlag Berlin Heidelberg} , \bibinfo{pages}{983--1000}.
\bibitem[{Treiber and Kesting(2017)}]{treiber2017intelligent}
\bibinfo{author}{Treiber, M.}, \bibinfo{author}{Kesting, A.},
  \bibinfo{year}{2017}.
\newblock \bibinfo{title}{The intelligent driver model with stochasticity-new
  insights into traffic flow oscillations}.
\newblock \bibinfo{journal}{Transportation research procedia}
  \bibinfo{volume}{23}, \bibinfo{pages}{174--187}.
\bibitem[{Vecerik et~al.(2017)Vecerik, Hester, Scholz, Wang, Pietquin, Piot,
  Heess, Roth{\"o}rl, Lampe and Riedmiller}]{vecerik2017leveraging}
\bibinfo{author}{Vecerik, M.}, \bibinfo{author}{Hester, T.},
  \bibinfo{author}{Scholz, J.}, \bibinfo{author}{Wang, F.},
  \bibinfo{author}{Pietquin, O.}, \bibinfo{author}{Piot, B.},
  \bibinfo{author}{Heess, N.}, \bibinfo{author}{Roth{\"o}rl, T.},
  \bibinfo{author}{Lampe, T.}, \bibinfo{author}{Riedmiller, M.},
  \bibinfo{year}{2017}.
\newblock \bibinfo{title}{Leveraging demonstrations for deep reinforcement
  learning on robotics problems with sparse rewards}.
\newblock \bibinfo{journal}{arXiv preprint arXiv:1707.08817} .
\bibitem[{Wang et~al.(2018)Wang, Chan and
  de~La~Fortelle}]{wang2018reinforcement}
\bibinfo{author}{Wang, P.}, \bibinfo{author}{Chan, C.Y.},
  \bibinfo{author}{de~La~Fortelle, A.}, \bibinfo{year}{2018}.
\newblock \bibinfo{title}{A reinforcement learning based approach for automated
  lane change maneuvers}, in: \bibinfo{booktitle}{2018 IEEE Intelligent
  Vehicles Symposium (IV)}, \bibinfo{organization}{IEEE}. pp.
  \bibinfo{pages}{1379--1384}.
\bibitem[{Wiedemann(1974)}]{Wiedemann.1974}
\bibinfo{author}{Wiedemann, R.}, \bibinfo{year}{1974}.
\newblock \bibinfo{title}{Simulation des Stra{\ss}enverkehrsflusses}.
\newblock \bibinfo{publisher}{{Institut f{\"u}r Verkehrswesen, University of
  Karlsruhe, Germany}}, \bibinfo{address}{Vol. 8 of Schriftenreihe des IfV}.

\end{thebibliography}





\newpage
\appendix

\section{Evaluation results of two-stage DDPG followers with different ratios}
\label{sec:appendix a}

\begin{figure}[htbp]
    \centering
    \includegraphics[width=\textwidth]{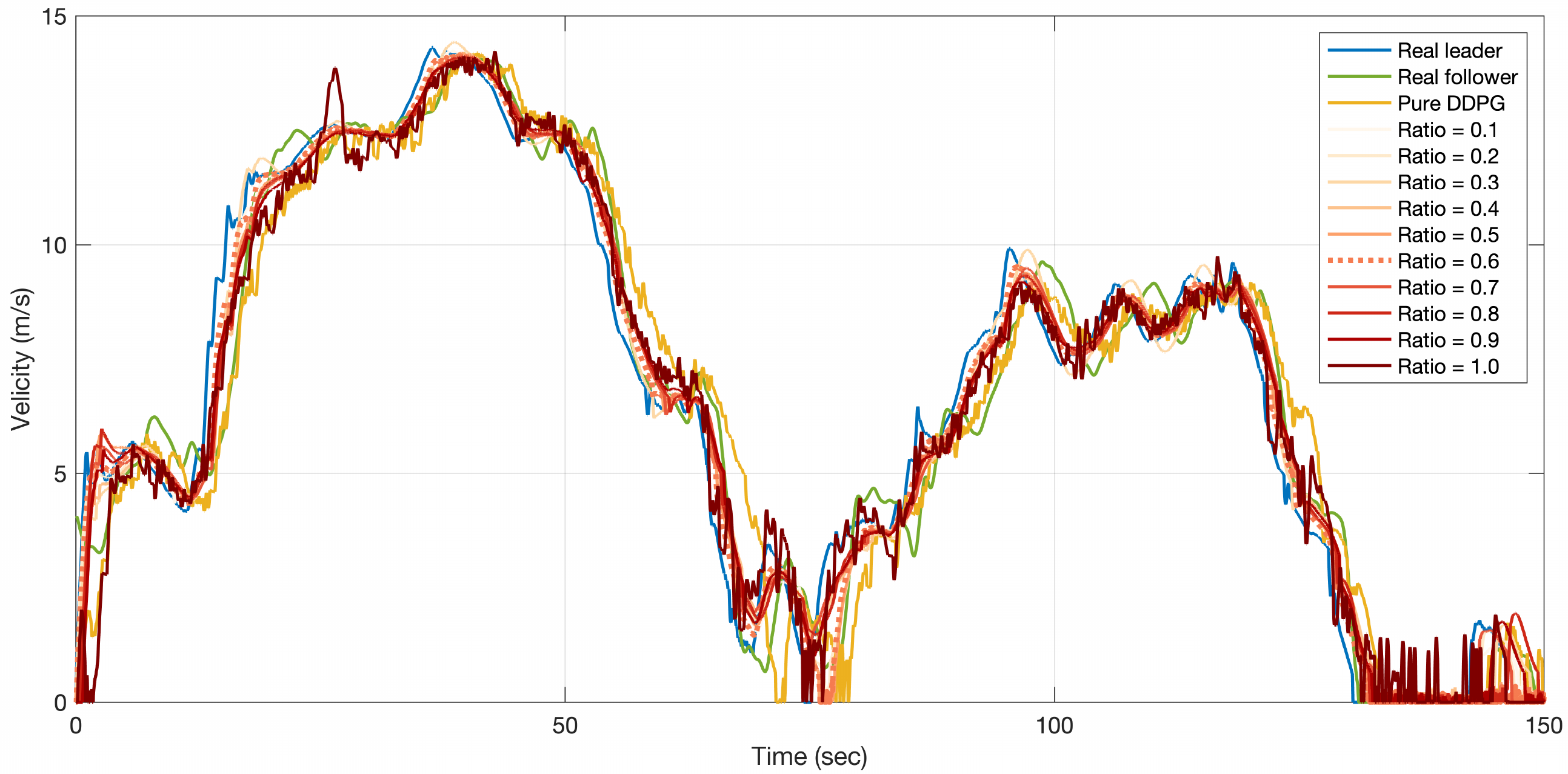}
    \caption{The evaluation results by following a real leader trajectory (blue) from Napoli datasets for two-stage DDPG followers with different utilization ratios of the real dataset.}
    \label{fig:different_velocity}
\end{figure}

\end{document}